%% file: main.tex
\documentclass[10pt,journal,compsoc]{IEEEtran}
\usepackage{times}
\usepackage{epsfig}
\usepackage{graphicx}
\usepackage{amsmath}
\usepackage{math}
\usepackage{amssymb}
\usepackage{esvect}

\usepackage{tabularx}
\usepackage{booktabs}
\usepackage{enumerate}
\usepackage{array,multirow}
\usepackage{enumitem}
\usepackage{caption}
\usepackage{subcaption}
\usepackage[export]{adjustbox}

\usepackage[table]{xcolor}%
\usepackage[pagebackref=true,breaklinks=true,letterpaper=true,bookmarks=false]{hyperref}

\newcommand{\xyh}[1]{\textcolor{black}{#1}}

\usepackage{review}
\setrevision{3}

\newcommand{\method}{TransCenter}
\newcommand{\methodlite}{TransCenter-Lite}
\newcommand{\PAR}[1]{\vskip4pt \noindent {\bf #1~}}

\newcommand{\dc}[1]{\textbf{\textsc{#1}}}

\ifCLASSOPTIONcompsoc
  \usepackage[nocompress]{cite}
\else
  \usepackage{cite}
\fi

\ifCLASSINFOpdf
\else
\fi

\begin{document}

\setcounter{page}{1}

\title{TransCenter: Transformers with Dense Representations for Multiple-Object Tracking}

\author{Yihong~Xu$^{*}$,
        Yutong~Ban$^{*}$,
        Guillaume~Delorme,
        Chuang~Gan,
        Daniela~Rus,~\IEEEmembership{Fellow,~IEEE, }~and~Xavier~Alameda-Pineda,~\IEEEmembership{Senior Member,~IEEE }%
\IEEEcompsocitemizethanks{\IEEEcompsocthanksitem Y. Xu, G. Delorme, X. Alameda-Pineda are with Inria Grenoble Rh\^one-Alpes, Montbonnot Saint-Martin, France. \protect\\
E-mail: \{firstname.lastname\}@inria.fr

\IEEEcompsocthanksitem Y. Ban, D. Rus are with Distributed Robotics Lab, CSAIL, Massachusetts Institute of Technology.\protect\\
E-mail: \{yban, rus\}@csail.mit.edu

\IEEEcompsocthanksitem C. Gan is with MIT-IBM Watson AI Lab.\protect\\
E-mail: ganchuang@csail.mit.edu
}%
\thanks{Y. Xu and Y. Ban contributed equally to this work.}}

\markboth{}%
{Xu \MakeLowercase{\textit{et al.}}: TransCenter: Transformers with Dense Representations for Multiple-Object Tracking}

\IEEEtitleabstractindextext{%
\begin{abstract}
Transformers have proven superior performance for a wide variety of tasks since they were introduced. In recent years, they have drawn attention from the vision community in tasks such as image classification and object detection. Despite this wave, an accurate and efficient multiple-object tracking (MOT) method based on transformers is yet to be designed. We argue that the direct application of a transformer architecture with quadratic complexity and insufficient noise-initialized sparse queries -- is not optimal for MOT. We propose~\method, a transformer-based MOT architecture with dense representations for accurately tracking all the objects while keeping a reasonable runtime. Methodologically, we propose the use of image-related dense detection queries and efficient sparse tracking queries produced by our carefully designed query learning networks (QLN). On one hand, the dense image-related detection queries allow us to infer targets' locations globally and robustly through dense heatmap outputs. On the other hand, the set of sparse tracking queries efficiently interacts with image features in our~\method\ Decoder to associate object positions through time. As a result,~\method~exhibits remarkable performance improvements and outperforms by a large margin the current state-of-the-art methods in two standard MOT benchmarks with two tracking settings (public/private). \method\ is also proven efficient and accurate by an extensive ablation study and comparisons to more naive alternatives and concurrent works. For scientific interest, the code is made publicly available at \url{https://github.com/yihongxu/transcenter}.
\end{abstract}

\begin{IEEEkeywords}
Multiple-Object Tracking, Efficient Transformer, Dense Image-Related Detection Queries, Sparse Tracking Queries.  %
\end{IEEEkeywords}}

\maketitle

\IEEEdisplaynontitleabstractindextext

\IEEEpeerreviewmaketitle

\input{intro}
\input{related_work}
\input{methodology}
\input{experiments}
\input{conclusion}

\appendices
\input{supple}
\ifCLASSOPTIONcompsoc
  \section*{Acknowledgments}
\else
  \section*{Acknowledgment}
\fi
Xavier Alameda-Pineda acknowledges funding from the ANR ML3RI project (ANR-19-CE33-0008-01) and the H2020 SPRING project (under GA \#871245).

{\small
\bibliographystyle{plain}
\bibliography{ref_short.bib}
}

\end{document}

%% file: intro.tex
\section{Introduction}
\xyh{The task of tracking multiple objects, usually understood as the simultaneous inference of the positions and identities of various objects/pedestrians (trajectories) in a visual scene recorded by one or more cameras, became a core problem in computer vision in the past years. Undoubtedly, the various multiple-object tracking (MOT) challenges and associated datasets~\cite{MOT16, MOTChallenge20}, helped foster research on this topic and provided a standard way to evaluate and monitor the performance of the methods proposed by many research teams worldwide.}

\xyh{ Recent progress in computer vision using transformers~\cite{vaswani2017attention} for tasks such as object detection~\cite{carion2020end,zhu2020deformable,lin2020detr}, person re-identification (Re-ID)~\cite{he2021transreid} or image super resolution~\cite{yang2020learning}, showed the benefit of the attention-based mechanism. Transformers are good at modeling simultaneously dependencies between different parts of the input and thus at making global decisions. These advantages fit perfectly the underlying challenges of MOT, where current methods often struggle when modeling the interaction between objects, especially in crowded scenes. We, therefore, are motivated to investigate the use of a transformer-based architecture for MOT, enabling global estimations when \addnote[R2Q4-1]{2}{finding} trajectories thus reducing missed or noisy predictions of trajectories.}
\begin{figure}[t!]
\centering
\begin{subfigure}{0.24\textwidth}
  \includegraphics[width=\textwidth]{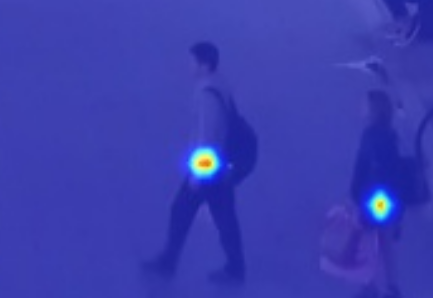}\vspace{-1.5mm}
  \caption{Detection}
  \label{fig:teaser_detection}
\end{subfigure}
\begin{subfigure}{0.24\textwidth}
  \includegraphics[width=\textwidth]{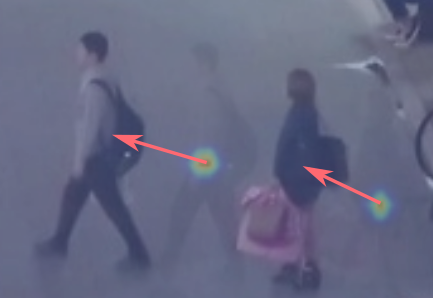}\vspace{-1.5mm}
  \caption{Tracking}
  \label{fig:teaser_tracking}
\end{subfigure}\hspace{1mm}%
\caption{\xyh{Via \method, we propose to tackle the MOT problem with transformers accurately and efficiently: the dense non-overlapping representations provide sufficient and accurate detections through \emph{dense image-size center heatmap} as shown in (a); The sparse tracking queries, obtained from features sampled within object positions at the previous time step, efficiently produce the \emph{sparse displacement vectors of objects} (shown in arrows plotted on the image originally with a gray background) from the previous to the current time step, as shown in (b).}}\vspace{-2mm}
\label{fig:teaserV1}
\end{figure}

Current MOT methods follow in principle the predominant tracking-by-detection paradigm where we first detect objects in the visual scene and associate them through time. The detection ability is critical for having a good MOT performance. To this end, MOT methods usually combine  probabilistic models~\cite{rezatofighi2015joint, ban2016tracking, Lin-TPAMI-2022, ban2019variational} or deep convolutional architectures~\cite{bergmann2019tracking, xu2020train,peng2020chained, Wang2021_GSDT, gan2019self,zhang2020multiplex, shan2020tracklets} with an integrated or external detector to predict bounding-box outputs. They are often based on overlapping predefined anchors that predict redundant outputs, which might create noisy detections, and thus need hand-crafted post-processing techniques such as non-maximum suppression (NMS), which might suppress correct detections. Instead, concurrent MOT methods~\cite{sun2020transtrack, meinhardt2021trackformer} built on recent object detectors like DETR~\cite{carion2020end} leverage transformers where sparse noise-initialized queries are employed, which perform cross-attention with encoded images to output one-to-one object-position predictions, thanks to the one-to-one assignment of objects and queries during training. However, using a fixed-number of sparse queries leads to a lack of detections when the visual scene becomes crowded, and the hyper-parameters of the model need to be readjusted. For these reasons, we argue that having dense but non-overlapping representations for detection is beneficial in MOT, especially in crowded scenes. 

Indeed, the image-size dense queries are naturally compatible with center heatmap predictions, and there exist center-based MOT methods~\cite{zhou2020tracking, zhang2020fairmot} that output dense object-center heatmap predictions directly related to the input image. Moreover, the center heatmap representations are pixel-related and thus inherit the one-to-one assignment of object centers and queries (pixels) without external assignment algorithms. However, \addnote[R3Q3]{2}{the locality of CNN architecture limits the networks to explore, as transformers do, the co-dependency of objects globally}. Therefore, we believe that the transformer-based approach with dense image-related (thus non-overlapping) representations is a better choice for building a powerful MOT method.

Designing such transformer-based MOT with dense image-related representations is far from evident. The main drawback is the computational efficiency related to the quadratic complexity in transformers w.r.t.\ the dense inputs. To overcome this problem, we introduce~\method, with powerful and efficient attention-based encoder and decoder, as well as the query generator.

For the encoder, the DETR~\cite{carion2020end} structure alleviates this issue by introducing a pre-feature extractor like ResNet-50 to extract lower-scale features before inputting the images to the transformers, but the CNN feature extractor itself contributes to the network complexity. Deformable transformers reduce significantly the attention complexity but the ResNet-50 feature extractor is still kept. Alternatively, recent efficient transformers like~\cite{wang2022pvt} discard the pre-feature extractor and input directly image patches, following the ViT structure~\cite{dosovitskiy2020image}. They also reduce the attention complexity in transformers by spatial-reduction attention, which makes building efficient transformer-based MOT possible. 

\xyh{In \method\ Decoder, the use of a deformable transformer indeed reduces the attention complexity. However, with our dense queries, the attention calculation remains heavy, which hinders computational efficiency. Alternatively, our image-size detection queries are generated from the lightweight query learning networks (QLN) that inputs image features from the encoder, and they are already globally related. Therefore, the heavy detection cross-attention with image features can be omitted. Furthermore, The tracking happens in the decoder where we search the current object positions with the known previous ones. With this prior positional information, we design the tracking queries to be sparse, which do not need to search every pixel for finding current object positions and significantly speed up the MOT method without losing accuracy.}

\addnote[R3Q2]{2}{To summarize, as roughly illustrated in Fig.~\ref{fig:teaserV1}, \method\ tackles the MOT problem with  image-related dense detection queries and sparse tracking queries.} It (i) introduces dense image-related queries to deal with miss detections from insufficient sparse queries or over detections in overlapping anchors of current MOT methods, yielding better MOT accuracy; (ii) solves the efficiency issue inherited in transformers with careful network designs, sparse tracking queries, and removal of useless external overheads, allowing \method\ to perform MOT efficiently. 
Overall, this work has the following contributions:
\xyh{\begin{itemize}
    \item We introduce \method, the first center-based transformer framework for MOT and among the first to show the benefits of using transformer-based architectures for MOT.
    \item We carefully explore different network structures to combine the transformer with center representations, specifically proposing \emph{dense image-related multi-scale representations} that are mutually correlated within the transformer attention and produce \emph{abundant but less noisy tracks, while keeping a good computational efficiency.}
    \item \addnote[R1Q5-3]{2}{We extensively compare with up-to-date online MOT tracking methods},~\method\ sets \emph{a new state-of-the-art} baseline both in MOT17~\cite{MOT16} (+4.0\% Multiple-Object Tracking Accuracy, MOTA) and MOT20~\cite{MOTChallenge20} (+18.8\% MOTA) by large margins, leading both MOT competitions by the time of our submission in the published literature. 
    \item Moreover, two more model options,~\method-Dual, which further boosts the performance in crowded scenes, and~\methodlite, enhancing the computational efficiency, are provided for different requirements in MOT applications.
\end{itemize}
}

%% file: related_work.tex
\section{Related Works}\label{sec:related_work}
\subsection{Multiple-Object Tracking}
In MOT literature, initial works \cite{ban2016tracking, rezatofighi2015joint, baisa2019online} focus on how to find the optimal associations between detections and tracks through probabilistic models while \cite{milan2017online} first formulates the problem as an end-to-end learning task with recurrent neural networks. Moreover, \cite{sadeghian2017tracking}  models the dynamics of objects by a recurrent network and further combines the dynamics with an interaction and an appearance branch. \cite{xu2020train} proposes a framework to directly use the standard evaluation measures MOTA and MOTP as loss functions to back-propagate the errors for an end-to-end tracking system. \cite{bergmann2019tracking} employs object detection methods for MOT by modeling the problem as a regression task. A person Re-ID network \cite{tang2017multiple, bergmann2019tracking} can be added at the second stage to boost the performance. However, it is still not optimal to treat the Re-ID as a secondary task.~\cite{zhang2020fairmot} further proposes a framework that treats the person detection and Re-ID tasks equally and~\cite{pang2021quasi} uses quasi-dense proposals from anchor-based detectors to learn Re-ID features for MOT in a dense manner.~\cite{shen2017fast} proposes a fast online MOT with detection refinement. Motion clues in MOT are also important since a well-designed motion model can compensate for missing detections due to occlusions. To this end,~\cite{yin2020unified} unifies the motion and affinity models for MOT.~\cite{liu2020multiple} leverages articulation information to perform MOT.~\cite{saleh2021probabilistic, han2020mat} focus on the motion-based interpolation based on probabilistic models.~\cite{wu2021track} unifies the segmentation and detection task in an MOT framework combining historic tracking information as a strong clue for the data association. Derived from the success in single-object tracking,~\cite{shuai2021siammot} employs siamese networks for MOT.~\cite{zhang2021bytetrack} simply leverages low-score detections to explore partially occluded objects. Moreover, traditional graphs are also used to model the positions of objects as nodes and the temporal connection of the objects as edges \cite{hornakova2020lifted, tang2017multiple, tang2015subgraph, keuper2016multi, tang2016multi}. The performance of those methods is further boosted by the recent rise of Graph Neural Networks (GNNs): hand-designed graphs are replaced by learnable GNNs \cite{weng2020gnn3dmot, weng2020joint, Wang2021_GSDT, papakis2020gcnnmatch, braso_2020_CVPR, he2021learnable} to model the complex interaction of the objects. 

Most of the above methods follow the tracking by detections/regression paradigm where a detector provides detections for tracking. The paradigm has proven state-of-the-art performance with the progress in object detectors. Unlike classic anchor-based detectors~\cite{ren2015faster, redmon2016you}, recent progress in keypoint-based detectors~\cite{law2018cornernet,zhou2019objects} exhibited better performance while discarding overlapping and manually-designed box anchors trying to cover all possible object shapes and positions. Built on keypoint-based detectors, \cite{zhou2020tracking}, \cite{zhang2020fairmot} and \cite{zheng2021improving} represent objects as centers in a heatmap then reason about all the objects jointly and associate them across adjacent frames with a tracking branch or Re-ID branch.

\subsection{Transformers in Multiple-Object Tracking}
Transformer is first proposed by \cite{vaswani2017attention} for machine translation and has shown its ability to handle long-term complex dependencies between entries in a sequence by using a multi-head attention mechanism. With its success in natural language processing, works in computer vision start to investigate transformers for various tasks, such as image recognition \cite{dosovitskiy2020image}, Person Re-ID \cite{he2021transreid}, realistic image generation \cite{jiang2021transgan}, super-resolution \cite{yang2020learning} and audio-visual learning~\cite{gan2020music,gan2020foley}.

Object detection with Transformer (DETR)~\cite{carion2020end} can be seen as an exploration and correlation task. It is an encoder-decoder structure where the encoder extracts the image information and the decoder finds the best correlation between the object queries and the encoded image features with an attention module. The attention module transforms the inputs into \emph{Query} ($Q$), \emph{Key} ($K$), and \emph{Value} ($V$) with fully-connected layers. Having $Q, K, V$, the attended features are calculated with the attention function~\cite{vaswani2017attention}:
\begin{equation}\label{eq:transformer}
    \text{Attention}(Q, K, V) = \text{Softmax}(\frac{QK^{T}}{\sqrt{h}}) V
\end{equation}
where $h$ is the hidden dimension of $Q, K,$ and $V$. The attention calculation suffers from heavy computational and memory complexities w.r.t the input size: the feature maps extracted from a ResNet-50~\cite{he2016deep} backbone are used to alleviate the problem. Deformable DETR~\cite{zhu2020deformable} further tackles the issue by proposing deformable attention inspired by~\cite{dai2017deformable}, drastically speeding up the convergence  (10$\times$) and reducing the complexity. The reduction of memory consumption allows in practice using multi-scale features to capture finer details, yielding better detection performance. However, the CNN backbone is still kept in Deformable DETR. This important overhead hinders the Transformer from being efficient. 
Alternatively, Pyramid Vision Transformer~\cite{wang2022pvt} (\emph{PVT}) extracts the visual features directly from the input images and the attention is calculated with efficient spatial-reduction attention (SRA). Precisely, PVT follows the ViT~\cite{dosovitskiy2020image} structure while the feature maps are gradually down-scaled with a patch embedding module (with convolutional layers and layer normalization). To reduce the quadratic complexity of Eq.~\ref{eq:transformer} w.r.t the dimension $d$ of $Q, K, V$, the SRA in PVT reduces beforehand the dimension of $K, V$ from $\Reel^{d \times h}$ to $\Reel^{\frac{d}{r}\times h}$  (with $r>1$, the scaling factor) and keeps the dimension of $Q$ unchanged. The complexity can be reduced by $r^2$ times while the dimension of the output attended features remains unchanged, boosting the efficiency of transformer attention modules. 

The use of transformers is still recent in MOT. Before transformers, some attempts with simple attention-based modules have been introduced for MOT. Specifically,~\cite{guo2021online} proposes a target and distractor-aware attention module to produce more reliable appearance embeddings, which also helps suppress detection drift and~\cite{wang2021multiple} proposes hand-designed spatial and temporal correlation modules to achieve long-range information similar to what transformers inherit. After the success in detection using transformers, two \emph{concurrent works} directly apply transformers on MOT based on the (deformable) DETR framework. First, Trackformer~\cite{meinhardt2021trackformer} builds directly from DETR~\cite{carion2020end} and is trained to propagate the queries through time. Second, Transtrack~\cite{sun2020transtrack} extends~\cite{zhu2020deformable} to MOT by adding a decoder that processes the features at $t-1$ to refine previous detection positions. Importantly, both methods stay in the DETR framework with sparse queries and extend it for tracking. However, recent literature~\cite{zhou2020tracking, zhang2020fairmot, zheng2021improving} also suggests that point-based tracking may be a better option for MOT while the use of pixel-level dense queries with transformers to predict dense heatmaps for MOT has never been studied. In addition, we question the direct transfer from DETR to MOT as concurrent works do~\cite{sun2020transtrack, meinhardt2021trackformer}. Indeed, the sparse queries without positional correlations might be problematic in two folds. Firstly, the insufficient number of queries could cause severe miss detections thus false negatives (FN) in tracking. Secondly, queries are highly overlapping, and simply increasing the number of non-positional-correlated queries may end up having many false detections and false positives (FP) in tracking. All of the above motivates us to investigate a better transformer-based MOT framework. We thus introduce~\method\, a methodology that takes existing drawbacks into account at the design level and achieves state-of-the-art performance.

%% file: methodology.tex
\begin{figure*}[ht!]
\centering\includegraphics[width=\linewidth]{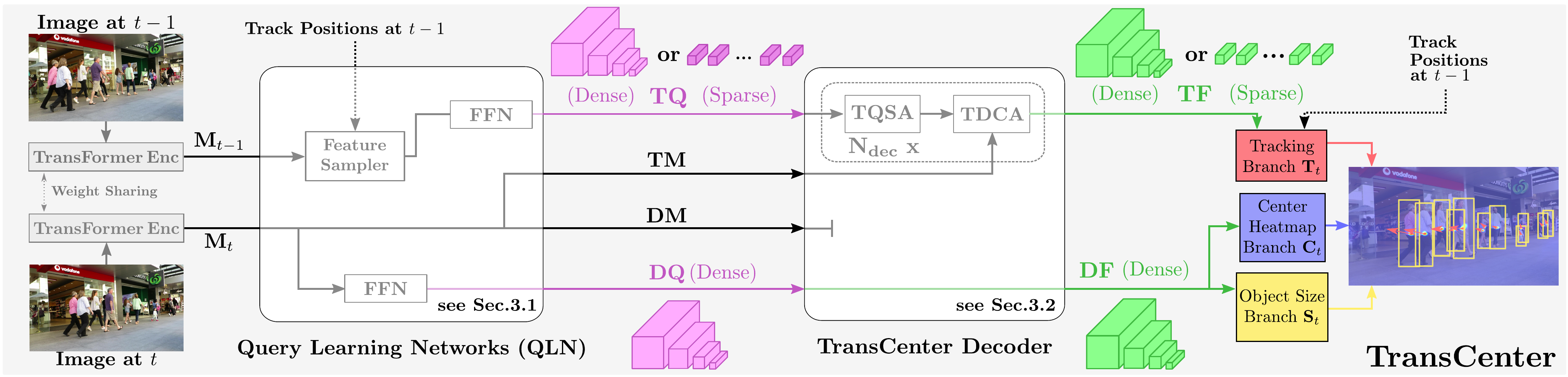}
  \caption{Generic pipeline of~\method\ and different variants: Images at $t$ and $t-1$ are fed to the transformer encoder (\emph{DETR-Encoder} or~\emph{PVT-Encoder}) to produce multi-scale memories $ \mathbf{M}_t$ and $ \mathbf{M}_{t-1}$ respectively. They are passed (together with track positions at $t-1$) to the \emph{Query Learning Networks (QLN)} operating in the feature's channel. QLN produce (1) dense pixel-level multi-scale detection queries--$ \mathbf{DQ}$, (2) detection memory--$ \mathbf{DM}$, (3) (sparse or dense) tracking queries--$ \mathbf{TQ}$, (4) tracking memory--$ \mathbf{TM}$. For associating objects through frames, the~\method\ Decoder performs cross attention between $ \mathbf{TQ}$ and $ \mathbf{TM}$, producing Tracking Features--$ \mathbf{TF}$. For detection, the~\method\ Decoder either calculates the cross attention between $ \mathbf{DQ}$ and $ \mathbf{DM}$ or directly outputs $ \mathbf{DQ}$ (in our efficient versions,~\method\ and \methodlite, see Sec.~\ref{sec:methodology}), resulting in Detection Features--$ \mathbf{DF}$ for the output branches, $ \mathbf{S}_t$ and $ \mathbf{C}_t$. $ \mathbf{TF}$, together with object positions at $t-1$ (sparse $\mathbf{TQ}$) or center heatmap $ \mathbf{C}_{t-1}$ (omitted in the figure for simplicity) and $\mathbf{DF}$ (dense $\mathbf{TQ}$), are used to estimate image center displacements $ \mathbf{T}_t$ indicating for each center its displacement in the adjacent frames (red arrows). We detail \emph{our choice} (\method) of QLN and \method\ Decoder structures in the figure. Other designs of QLN and \method\ Decoder are detailed in Fig.~\ref{fig:qln} and Fig.~\ref{fig:transformerdecoder}. Arrows with a dotted line are only necessary for models with sparse $ \mathbf{TQ}$.}
  
\label{fig:pipeline}
\vspace{-2ex}
\end{figure*}
\section{\method{}}\label{sec:methodology}
\method{} tackles the detection and temporal association in MOT accurately and efficiently. Different from concurrent transformer-based MOT methods, \method{} questions the use of sparse queries without image correlations (i.e.~noise initialized), and explores the use of image-related dense queries, producing dense representations using transformers. To that aim, we introduce the query learning networks (QLN) responsible for converting the outputs of the encoder into the inputs of the \method\ Decoder. Different possible architectures for QLN and \method\ Decoder are proposed, and the choice is made based on both the accuracy and the efficiency aspects. 

While exploiting dense representations from dense queries\footnote{See visualization in Supplementary Material Sec.~B.2.}, can help to sufficiently detect the objects, especially in crowded scenes, the design of dense queries is not trivial. Notably, the quadratic increase of calculation complexity in transformers should be solved, and the noise tracks from randomly initialized dense queries should be addressed. To this end, we propose image-related dense queries, which have three prominent advantages: \addnote[R4Q6]{2}{(1) the queries are multi-scale and exploit the multi-resolution structure of the encoder, allowing for very small targets to be captured by those queries; (2) image-related dense detection queries also make the network more flexible. The number of queries grows with the input resolution. No fixed hyper-parameters like in~\cite{sun2020transtrack} need to be adjusted to re-train the model, which depends on the density of objects in the scene; (3) the query-pixel correspondence discards the time-consuming Hungarian matching~\cite{Kuhn55thehungarian} for the query-ground-truth object association. Up to our knowledge, we are the first to explore the use of image-related dense detection queries that scale with the input image size. Meanwhile, we solve the efficiency issue through a careful network design that appropriately handles the dense representation of queries, which provides an accurate and efficient MOT method.}

A generic pipeline of \method\ is illustrated in Fig.~\ref{fig:pipeline}. RGB images at $t$ and $t-1$ are input to the weight-shared \addnote[R3Q5]{2}{transformer encoder} from which dense multi-scale attended features are obtained, namely memories $ \mathbf{M}_{t}$ and $ \mathbf{M}_{t-1}$. They are the inputs of the QLN. \addnote[R1Q8]{2}{QLN} produces two sets of output pairs, \addnote[R3Q4]{2}{detection} queries ($ \mathbf{DQ}$) and memory ($ \mathbf{DM}$) for detecting the objects at time $t$, and tracking queries ($ \mathbf{TQ}$) and memory ($ \mathbf{TM}$) for associating the objects at $t$ with previous time step $t-1$. Furthermore, \method\ Decoder, leveraging the deformable transformer~\cite{zhu2020deformable}, is used to correlate $ \mathbf{DQ}$/$ \mathbf{TQ}$ with $ \mathbf{DM}$/$ \mathbf{TM}$. To elaborate,  $ \mathbf{TQ}$ interacts with $ \mathbf{TM}$ in the cross-attention module of the \method\ Decoder, resulting in the tracking features ($\mathbf{TF}$). Similarly, the detection features ($\mathbf{DF}$) are the output of the cross-attention between $ \mathbf{DQ}$ and $ \mathbf{DM}$. To produce the output dense representations, $ \mathbf{DF}$ are used to estimate the object size $ \mathbf{S}_t$ and the center heatmap $ \mathbf{C}_t$. Meanwhile $ \mathbf{TF}$ are used to estimate the tracking displacement $ \mathbf{T}_t$.

One can argue that the downside of using dense queries is the associated higher memory consumption and lower computational efficiency. One drawback with previous or concurrent approaches is the use of the deformable DETR encoder including the CNN feature extractor ResNet-50~\cite{he2016deep}, which significantly slows down the feature extraction. Instead, \addnote[R4Q5]{2}{\method\ leverages PVTv2~\cite{wang2022pvt} as its encoder, the so-called \emph{PVT-Encoder}. The reasons are three-fold: (1) it discards the ResNet-50 backbone and uses efficient attention heads~\cite{wang2022pvt}, reducing significantly the network complexity; (2) it has flexible scalability by modifying the feature dimension $h$ and block structures. Specifically, we use B0~(denoted as \emph{PVT-Lite}) in \methodlite\ and B2~(\emph{PVT-Encoder}) for~\method\ and~\method-Dual, see details in~\cite{wang2022pvt} and Sec.\ref{subsec:implementdetails}}; (3) its feature pyramid structure is suitable for building dense pixel-level multi-scale queries.

Once the \addnote[R3Q6]{2}{transformer encoder} extracts the dense memory representations $\mathbf{M}_{t}$ and $ \mathbf{M}_{t-1}$, they are passed to QLN and then to the \method\ Decoder. We carefully search the design choices of QLN (see Sec.~\ref{subsec:dlq}) and \method\ Decoder (see Sec.~\ref{subsec:dualdecoders}), and select the best model based on the efficiency and accuracy. In particular, we demonstrate that $ \mathbf{TQ}$ can be sparse \footnote{See visualization in Supplementary Material Sec.~B.3.}, different from $ \mathbf{DQ}$, since we have the prior information of object positions at the previous time step that helps to search their corresponding positions at the current time step. Therefore, it is not necessary to search every pixel for this aim. Consequently, the discretization of tracking queries based on previous object positions (roughly from 14k to less than 500 depending on the number of objects) can significantly speed up the tracking attention calculation in the \method\ Decoder. 

Regarding the detection, the cross attention module between the dense detection queries ($ \mathbf{DQ}$) and the detection memory ($ \mathbf{DM}$) is beneficial in terms of performance but at the cost of significant computational loads. \addnote[R4Q5tier]{2}{We solve this by studying the impact of the detection cross-attention on the computational efficiency and the accuracy. We introduce two variants of the proposed method, \emph{\method-Dual} and \emph{\methodlite}. The former shares the same structure as \method\ but having DDCA for detection in the decoder, as detailed in Sec.~\ref{subsec:dualdecoders}; The latter is a lighter version of \method\  with a lighter encoder (PVT-Lite).} 

In the following sections, we detail the design choices of the QLN and the \method\ Decoder, and provide the details of the final output branches (see Sec.~\ref{subsec:branches}) as well as the training losses (see Sec.~\ref{subsec:losses}).

\subsection{QLN: Query Learning Networks}~\label{subsec:dlq}
\definecolor{Dq}{RGB}{3,102,173}
\definecolor{Mt}{RGB}{0,159,54}
\definecolor{SE-}{RGB}{134, 0, 0}
\begin{figure}[ht]
\captionsetup[subfigure]{labelformat=empty}
\centering
\begin{subfigure}{0.24\textwidth}
  \includegraphics[width=\textwidth]{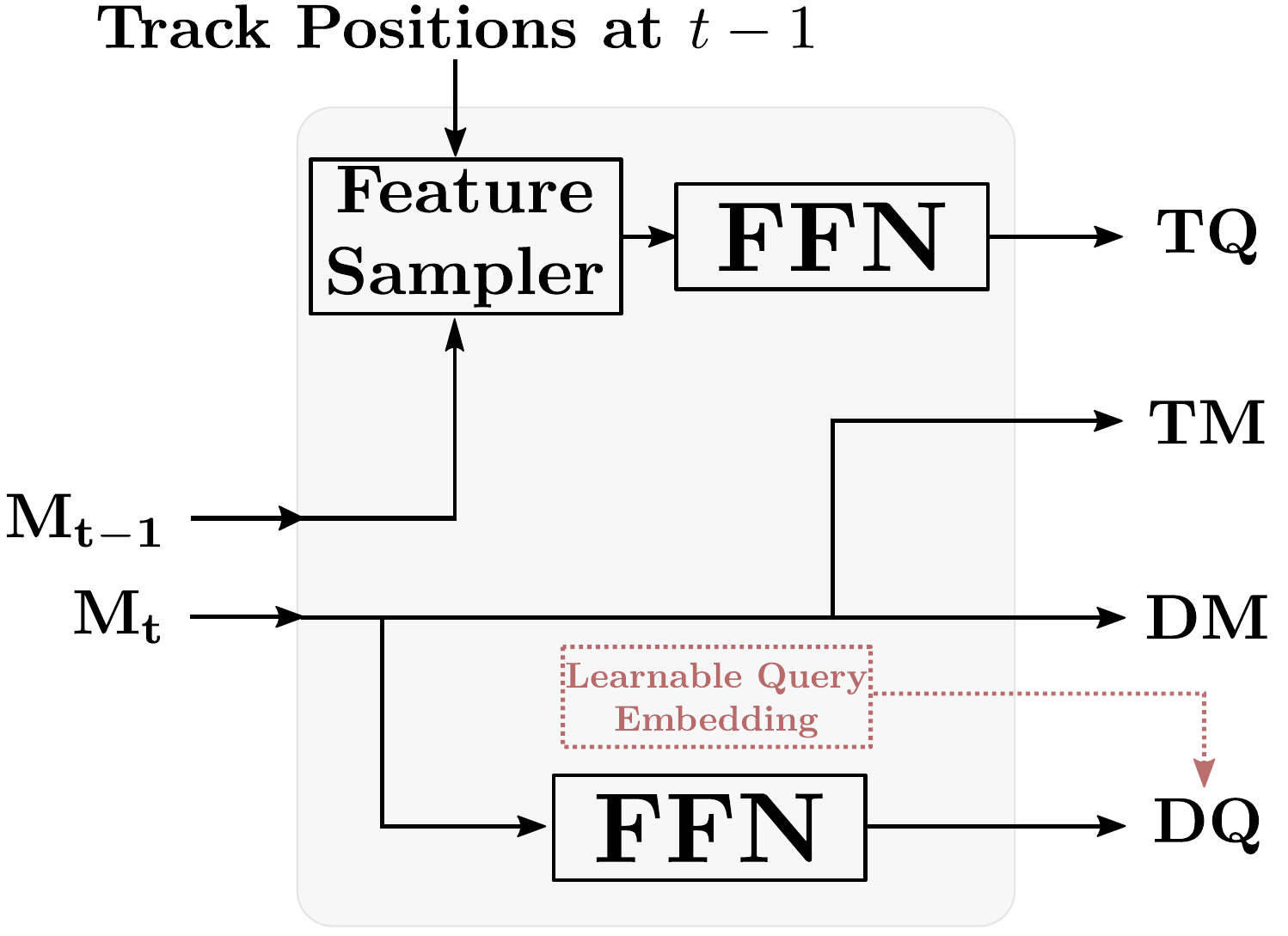}\vspace{-1.5mm}
  \caption{(a) QLN$_{S-}$ and \textcolor{SE-}{QLN$_{SE-}$}}
  \label{fig:qln_a_proposed}
\end{subfigure}
\begin{subfigure}{0.24\textwidth}
  \includegraphics[width=\textwidth]{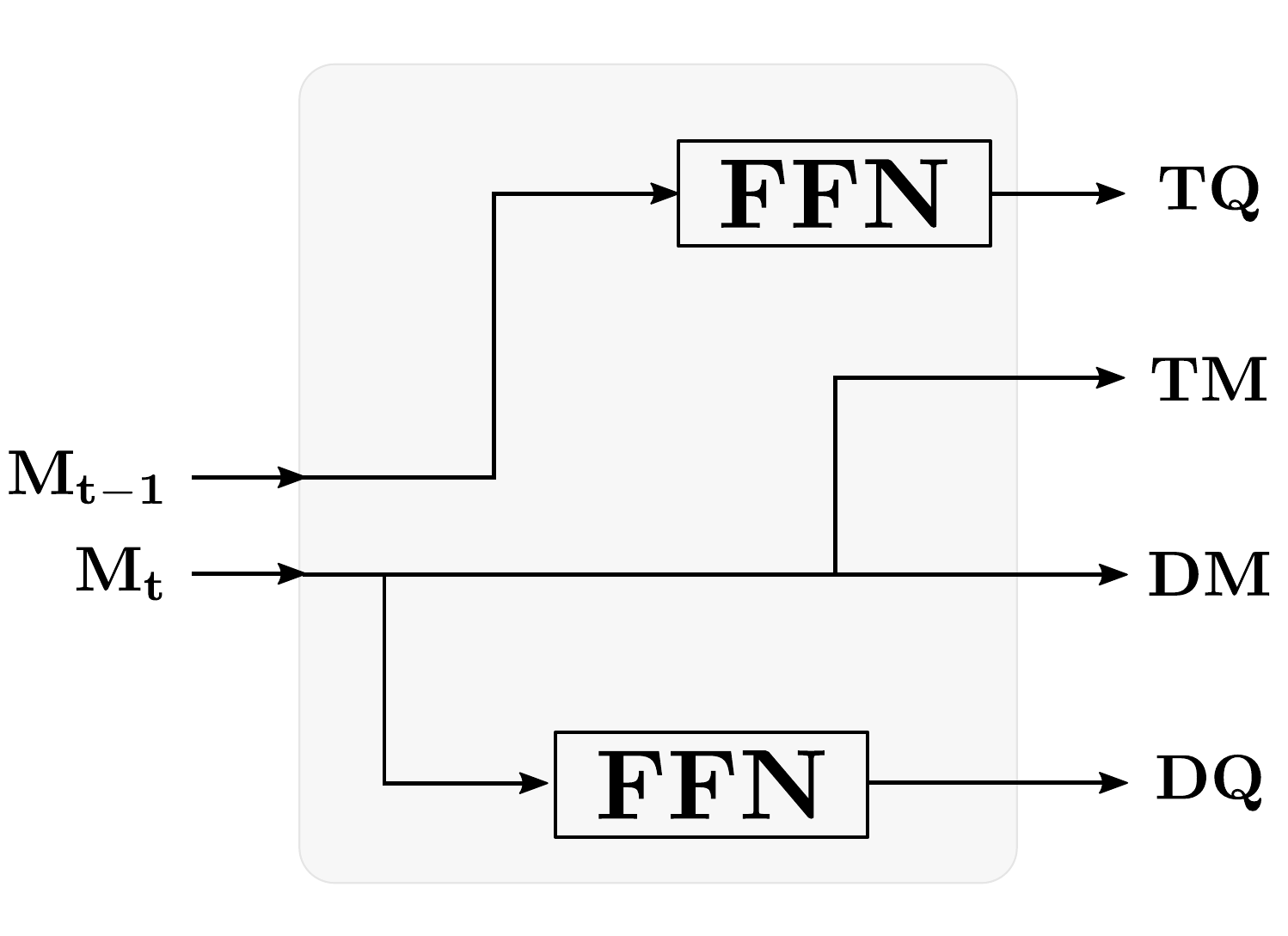}\vspace{-1.5mm}
  \caption{(b) QLN$_{D-}$}
  \label{fig:qln_b_old_trctr_mt-1}
\end{subfigure}\hspace{1mm}%
\begin{subfigure}{0.24\textwidth}
  \includegraphics[width=\textwidth]{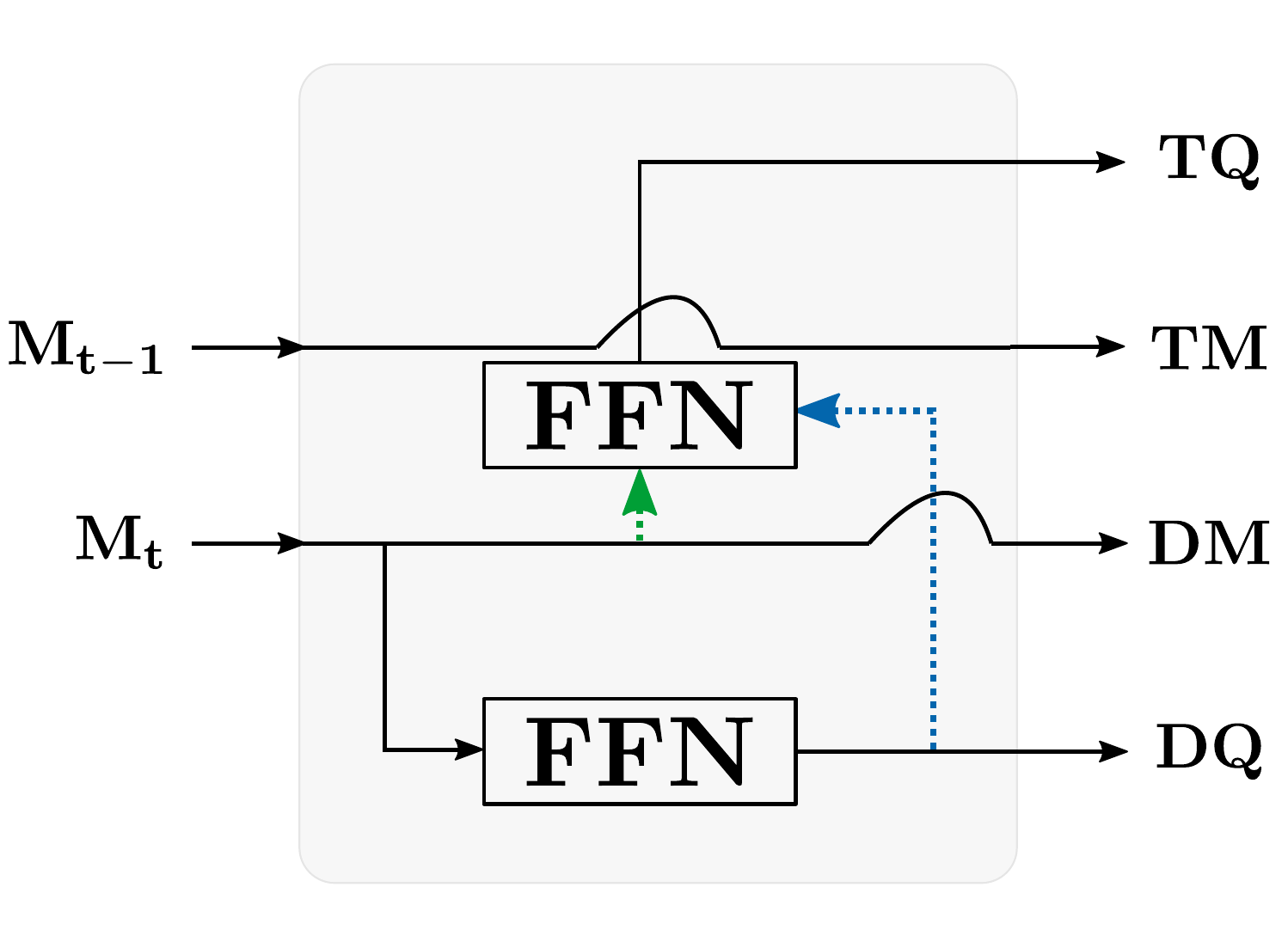}\vspace{-1.5mm}
  \caption{(c) QLN$_{D}$}
  \label{fig:qln_b_old_trctr_mt_dq}
\end{subfigure}
\begin{subfigure}{0.24\textwidth}
  \includegraphics[width=\textwidth]{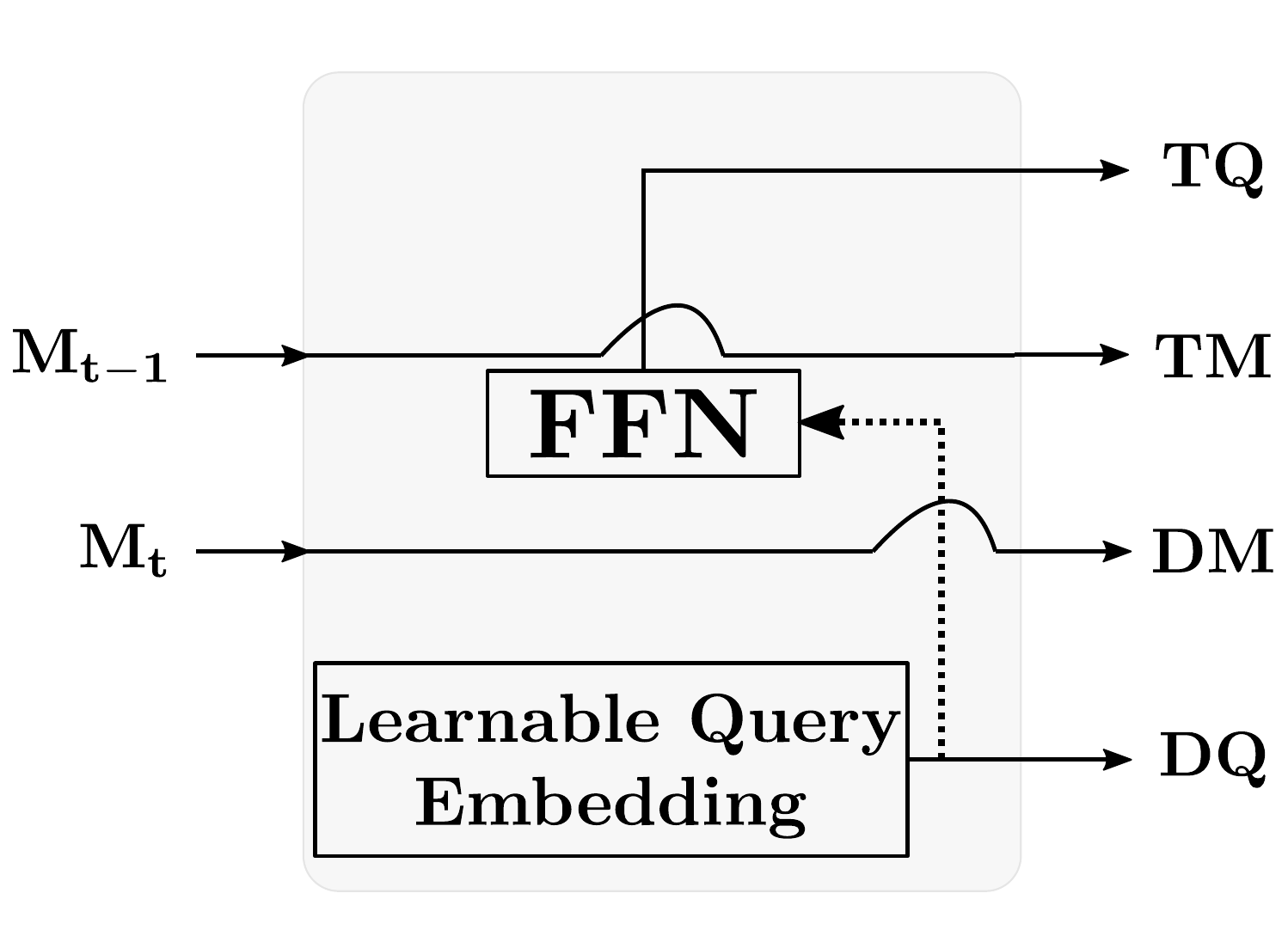}\vspace{-1.5mm}
  \caption{(d) QLN$_{E}$}
  \label{fig:qln_d_old_trctr_naive}
\end{subfigure}
\caption{Query Learning Networks (QLN): \method\ uses QLN$_{S-}$ (our choice) as its query learning network, producing sparse tracking queries by sampling prior object features from $\mathbf{M}_{t-1}$. Different structures of QLN are studied such as \textcolor{SE-}{QLN$_{SE-}$}, QLN$_{D-}$, QLN$_{D}$ (QLN$_{M_t}$ in green arrow and QLN$_{DQ}$ in blue arrow), and QLN$_{E}$, detailed in Sec.~\ref{subsec:dlq}. Best seen in color.}\label{fig:qln}
\end{figure}

QLN are networks that relate the queries and the memories. In our design, (1) $ \mathbf{DQ}$ are dense and image-related that discover object positions precisely and abundantly. (2) Different from $\mathbf{DQ}$, \addnote[R1Q3-1]{1}{$ \mathbf{TQ}$ are sparse and aim at finding object displacements between two different frames, and thus $ \mathbf{TQ}$ and $ \mathbf{TM}$ \textit{should be produced by input features from different time steps}.} 

Based on these attributes, we design the chosen QLN$_{S-}$\footnote{"S" means sparse, "$-$" means that the features are sampled from $t-1$. It is counter-intuitive to have QLN$_{S}$ because at time $t$, we know neither the number of tracked objects (tracks) at $t$ nor their positions, we cannot thus sample features with track positions.} that produces $ \mathbf{DQ}$ by passing $ \mathbf{M}_t$ (attended features from image $t$) through a FFN (feed-forward network with fully-connected layers). For $ \mathbf{TQ}$, QLN$_{S-}$ samples object features with a feature sampler using bilinear interpolation from $\mathbf{M}_{t-1}$ using object positions at $t-1$, while outputting $ \mathbf{TM}$ from features at a different time step, $\mathbf{M}_{t}$. Moreover, \addnote[R5Q3]{2}{different possible variants of QLN are visualized in Fig.~\ref{fig:qln}} and are summarized as follows:
\begin{enumerate}
    \item QLN$_{D-}$: QLN$_{S-}$ with \emph{dense} (with the subscript "D") $\mathbf{TQ}$ without the feature sampler.
    
    \item QLN$_{M_t}$: QLN$_{D-}$ with dense $\mathbf{TQ}$ from $\mathbf{M}_t$ and $\mathbf{TM}$ from $\mathbf{M}_{t-1}$.
    
    \item QLN$_{DQ}$: QLN$_{D-}$ with dense $\mathbf{TQ}$ from $\mathbf{DQ}$ and $\mathbf{TM}$ from $\mathbf{M}_{t-1}$.
    
    \item QLN$_{E}$: QLN$_{DQ}$ with noise-initialized learnable embeddings (with the subscript "E").
    
    \item QLN$_{SE-}$: QLN$_{S-}$ with noise-initialized learnable embeddings for detection.
\end{enumerate}

In Sec.~\ref{subsec:ablation}, different QLN are carefully designed and ablated to produce dense detection queries relative to the input image and sparse tracking queries for accurate and efficient MOT with transformers.
\definecolor{gray}{RGB}{153,153,153}
\begin{figure}[t]
\centering
 \includegraphics[width=.7\linewidth]{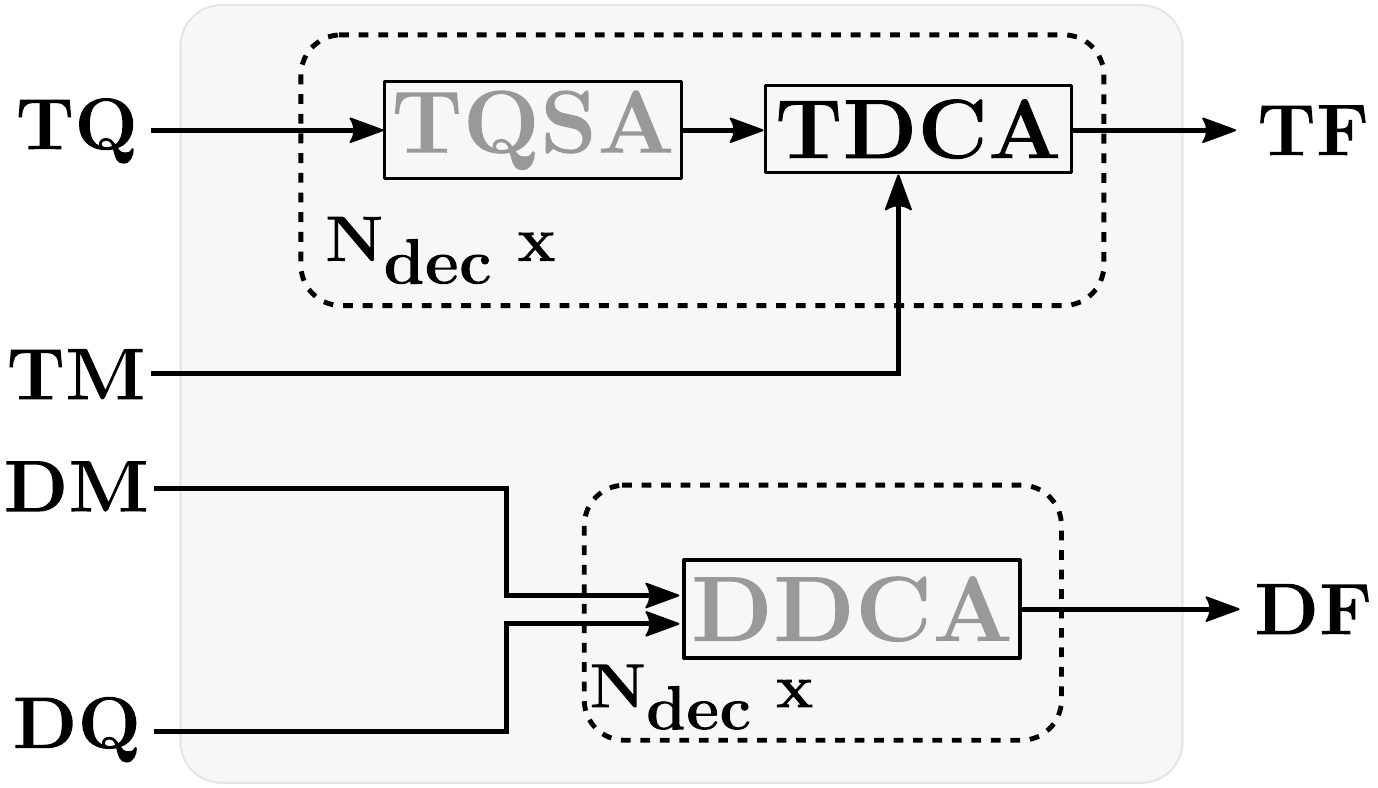}
\caption{\method\ Decoder is used to handle tracking queries $ \mathbf{TQ}$ and detection queries $ \mathbf{DQ}$. The detection attention correlates $ \mathbf{DQ}$ and $ \mathbf{DM}$ with the attention modules to detect objects. The tracking attention correlates $ \mathbf{TQ}$ and $ \mathbf{TM}$ to learn the displacements of the tracked objects until $t-1$ between different frames. \method\ Decoder has three main modules TQSA, DDCA, and TDCA (detailed in Sec.~\ref{subsec:dualdecoders}). Different versions of \method\ Decoder depending on discarding the DDCA or not, are denoted as \emph{Single} or \emph{Dual} decoder respectively. Also, an extra prefix "TQSA-" is added if the decoder has TQSA\protect\footnotemark.
\method\ uses \emph{TQSA-Single} considering the efficiency-accuracy tradeoff. The choice is based on the ablation of the aforementioned variants in Sec.~\ref{subsec:ablation}. $\textbf{N}_{dec}$ is the number of decoder layers.}\label{fig:transformerdecoder}
\end{figure}\footnotetext{We note that discarding TDCA is impossible since the tracking queries at $t-1$ or $t$ should interact with $\mathbf{TM}$ for searching objects at $t$ or $t-1$.}

\subsection{\method\ Decoder}~\label{subsec:dualdecoders}
To successfully find object trajectories, a MOT method should not only detect the objects but also associate them across frames. To do so,~\method\ Decoder tackles in parallel the two sub-tasks: detection and object temporal association (tracking). Concretely, \method\ Decoder consists of Tracking Deformable Cross-Attention (TDCA), and Detection Deformable Cross-Attention (DDCA) modules. Moreover, Tracking Query Self-Attention (TQSA) module is introduced to enhance the interactions among sparse tracking queries through a multi-head self-attention, knowing that the overhead is acceptable because the tracking queries are sparse in~\method. TDCA calculates cross attention between tracking queries and memories ($ \mathbf{TQ}$ and $ \mathbf{TM}$), resulting in tracking features ($ \mathbf{TF}$). Analogously, DDCA calculates cross attention between detection queries and memories ($ \mathbf{DQ}$ and $ \mathbf{DM}$), producing detection features ($ \mathbf{DF}$). Deformable cross attention module~\cite{zhu2020deformable} with linear complexity w.r.t.~input size is used.

From the efficiency perspective, the use of the multi-head attention modules in traditional transformers~\cite{vaswani2017attention} like DETR~\cite{carion2020end} implies a complexity growth that is quadratic with the input size. Of course, this is undesirable and would limit the scalability and usability of the method. \addnote[R2Q1-1]{1}{To mitigate this, \addnote[R4Q5bis]{2}{we resort to the deformable multi-head attention~\cite{zhu2020deformable}--Deformable Cross-Attention (DCA), where the queries are input to produce sampling offsets for displacing the input reference points. The reference points are from either the track position at $t-1$ for tracking in TDCA or the pixel coordinates of the dense queries for detection in DDCA}\footnote{The reference points are omitted in the figures for simplicity.}. The displaced coordinates are used to locate and sample features in $ \mathbf{DM}$ or $ \mathbf{TM}$. The input queries also produce in DCA the attention weights for merging sampled features.} However, the cost of calculating the cross attention between $ \mathbf{DQ}$ and $ \mathbf{DM}$ is still not negligible because of their multi-scale image resolutions. To solve this, we demonstrate in Sec.~\ref{subsec:ablation} (\emph{Single} v.s.~\emph{Dual} decoder) that it is possible to output directly $ \mathbf{DQ}$ as $ \mathbf{DF}$ for output-branch predictions, with an acceptable loss of accuracy as expected. In addition, under this \textit{sparse} nature of $ \mathbf{TQ}$ in \method, we enhance the interactions among queries by adding lightweight TQSA before TDCA.

To conclude, we choose to use a \emph{TQSA-Single} \method\ Decoder for the cross attention of $ \mathbf{TQ}$ and $ \mathbf{TM}$ while we directly use $ \mathbf{DQ}$ for the output branches. This is possible thanks to the \emph{sparse} $ \mathbf{TQ}$ and \emph{dense} $ \mathbf{DQ}$, which yields a good balance between computational efficiency and accuracy, The overall Decoder design is illustrated in Fig.~\ref{fig:transformerdecoder} and the comparison between different design choices can be found in Sec.~\ref{subsec:ablation}.

\begin{figure}[t]
\centering
\includegraphics[width=\linewidth]{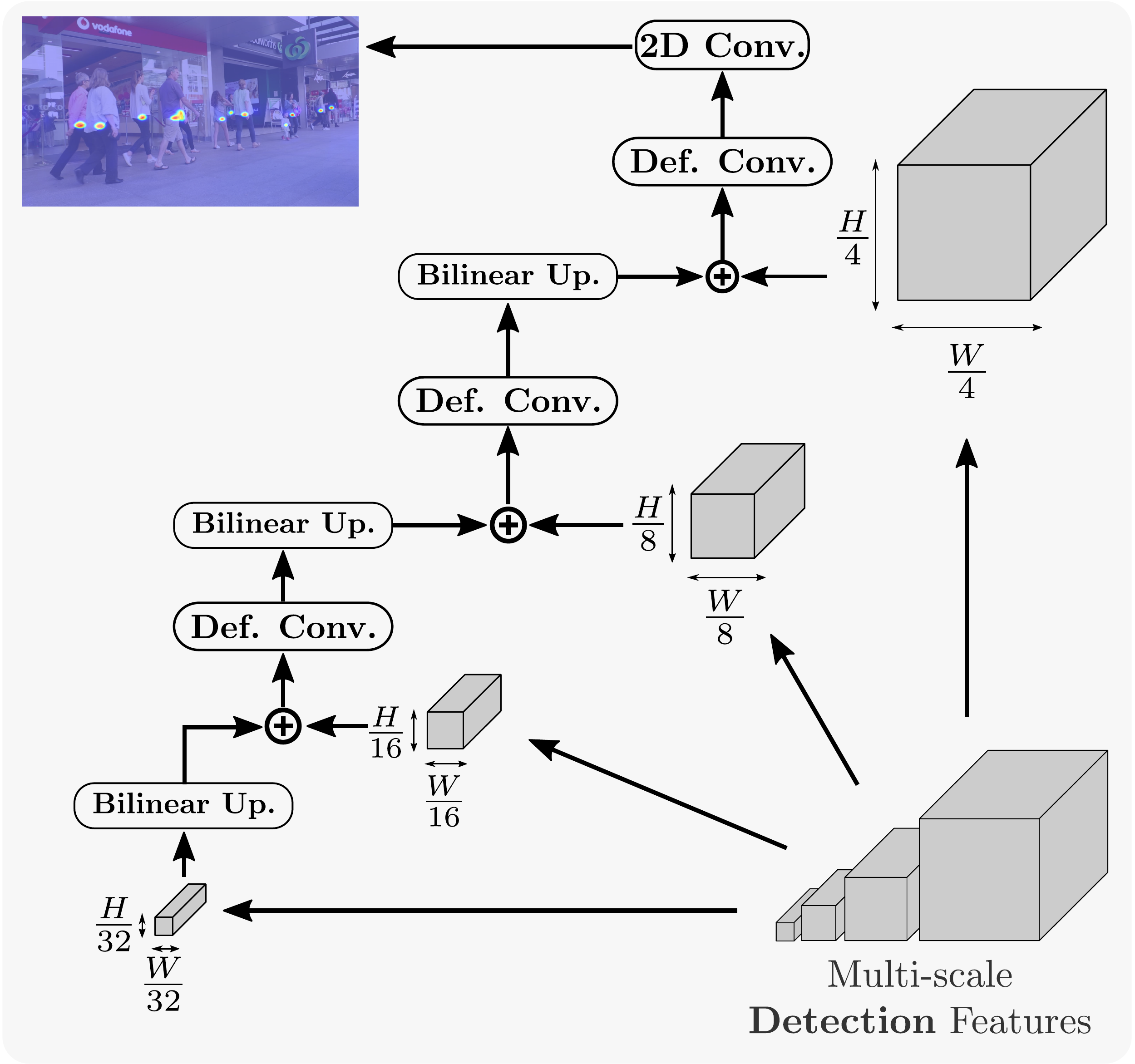}
\caption{Overview of the center heatmap branch. The multi-scale detection features are up-scaled (bilinear up.) and merged via a series of deformable convolutions (Def. Conv., the ReLU activation is omitted for simplicity)~\cite{dai2017deformable}, into the output center heatmap. A similar strategy is followed for the object size and the tracking branches.}
\label{fig:dla}
\end{figure}
\subsection{The Center, the Size, and the Tracking Branches}
\label{subsec:branches}
Given $ \mathbf{DF}$ and $ \mathbf{TF}$ from the~\method\ Decoder, we use them as the input of different branches to output object center heatmap $ \mathbf{C}_t$, their box size $ \mathbf{S}_t$ and the tracking displacements $ \mathbf{T}_t$. $ \mathbf{DF}$ contain feature maps of four different resolutions, namely $1/32, 1/16, 1/8$, and $1/4$ of the input image resolution. For the center heatmap and the object size, the feature maps at different resolutions are combined using deformable convolutions~\cite{dai2017deformable} and bilinear interpolation up-sampling, following the architecture shown in Fig.~\ref{fig:dla}. They are up-sampled into a feature map of $1/4$ of the input resolution and then input to $ \mathbf{C}_t\in[0,1]^{H/4\times W/4}$ and $ \mathbf{S}_t\in\mathbb{R}^{H/4\times W/4\times 2}$. $H$ and $W$ are the input image height and width, respectively, and the two channels of $ \mathbf{S}_t$ encode the object size in width and height. 

Regarding the tracking branch, the tracking features $ \mathbf{TF}$ are sparse having the same size as the number of tracks at $t-1$. One tracking query feature corresponds to one track at $t-1$. $ \mathbf{TF}$, together with object positions at $t-1$ ( in the case of sparse $\mathbf{TQ}$) or center heatmap $ \mathbf{C}_{t-1}$ and $\mathbf{DF}$ (dense $\mathbf{TQ}$), are input to two fully-connected layers with ReLU activation. The latter layers predict the horizontal and vertical displacements $ \mathbf{T}_t$ of objects at $t-1$ in the adjacent frames.

\subsection{Model Training}\label{subsec:losses}
The model training is achieved by jointly learning a 2D classification task for the object center heatmap and a regression task for the object size and tracking displacements, covering the branches of \method. For the sake of clarity, in this section, we will drop the time index $t$.
\PAR{Center Focal Loss.} To train the center branch, we need first to build the ground-truth heatmap response $ \mathbf{C}^{*}\in [0,1]^{H/4 \times W/4}$. As done in~\cite{zhou2020tracking}, we construct $ \mathbf{C}^{*}$ by considering the maximum response of a set of Gaussian kernels centered at each of the $K>0$ ground-truth object centers. More formally, for every pixel position $(x,y)$ the ground-truth heatmap response is computed as:
\begin{equation}\label{eq:gaussian}
     \mathbf{C}^*_{xy} = \max_{k=1,\ldots,K} G((x,y),(x_{k},y_{k});\sigma),
\end{equation}
where $(x_{k},y_{k})$ is the ground-truth object center, and $G(\cdot,\cdot;\sigma)$ is the Gaussian kernel with spread $\sigma$. In our case, $\sigma$ is proportional to the object's size, as described in~\cite{law2018cornernet}. %
Given the ground-truth $ \mathbf{C}^*$ and the inferred $ \mathbf{C}$ center heatmaps, the center focal loss, $L_{\textsc{c}}$ is formulated as:
\begin{equation}
        L_{\textsc{c}}= \frac{1}{K}\sum_{xy}
        \begin{cases}
         (1- \mathbf{C}_{xy})^{\alpha}\log( \mathbf{C}_{xy}) &   \mathbf{C}_{xy}^* = 1,\\
         (1- \mathbf{C}^*_{xy})^{\beta}( \mathbf{C}_{xy})^{\alpha}\log(1- \mathbf{C}_{xy})& \text{otherwise}.
        \end{cases}
\label{eq:2dfocal}
\end{equation}
where the scaling factors are $\alpha=2$ and $\beta=4$, see~\cite{zhang2020fairmot}.
\PAR{Sparse Regression Loss.} The values of $ \mathbf{S}$ is supervised only on the locations where object centers are present, i.e.~$C_{xy}^* = 1$ using a $L_1$ loss:
\begin{equation}\label{eq:regloss}
        L_{\textsc{s}} = \frac{1}{K}\sum_{xy}
        \begin{cases}
        \left \|   \mathbf{S}_{xy} -  \mathbf{S}^{*}_{xy} \right \|_1 &  \mathbf{C}_{xy}^* = 1,\\
        0 & \text{otherwise}.
        \end{cases}\\
\end{equation}
The formulation of $L_{\textsc{t}}$ for $ \mathbf{T}$ is analogous to $L_{\textsc{s}}$ but using the tracking output and ground-truth displacement, instead of the object size. To complete the sparsity of $L_{\textsc{s}}$, we add an extra $L_1$ regression loss, denoted as $L_{R}$ with the bounding boxes computed from $ \mathbf{S_{t}}$ and ground-truth centers. To summarize, the overall loss is formulated as the weighted sum of all the losses, where the weights are chosen according to the numeric scale of each loss:
\begin{equation}\label{eq:loss}
    L = L_{\textsc{c}} + \lambda_{\textsc{s}}L_{\textsc{s}} + \lambda_{\textsc{t}}L_{\textsc{t}} + \lambda_{\textsc{r}}L_{\textsc{r}}
\end{equation}

%% file: experiments.tex
\section{Experimental Evaluation}\label{sec:exp}
\input{implementation_details}
\input{tab/merge_results_SOTA_MOT17}
\subsection{Protocol}
\PAR{Datasets and Detections.} We use the standard split of the MOT17~\cite{MOT16} and MOT20~\cite{MOTChallenge20} datasets and the testset evaluation is obtained by submitting the results to the MOTChallenge website. The MOT17 testset contains 2,355 trajectories distributed in 17,757 frames. MOT20 testset contains 1,501 trajectories within only 4,479 frames, which leads to a much more challenging crowded-scene setting. We evaluate \method\ both under public and private detections. When using public detections, we limit the maximum number of birth candidates at each frame to the number of public detections per frame, as in~\cite{zhou2020tracking,meinhardt2021trackformer}. The selected birth candidates are those closest to the public detections with IOU larger than 0. When using private detections, there are no constraints, and the detections depend only on the network's detection capacity, the use of external detectors, and more importantly, the use of extra training data. For this reason, we regroup the results by the use of extra training datasets as detailed in the following. In addition, we evaluate our \method\ on the KITTI dataset under the autonomous driving setting. KITTI dataset contains annotations of cars and pedestrians in 21 and 29 video sequences in the training and test sets, respectively. For the results of KITTI dataset, we use also \cite{thomas_pandikow_kim_stanley_grieve_2021} as extra data.\input{tab/merge_results_SOTA_MOT20}
\PAR{Extra Training Data.} To fairly compare with the \addnote[R3Q5-1]{1}{state-of-the-art} methods, we denote the extra data used
to train each method, including several pre-prints listed in the MOTChallenge leaderboard, which are marked with * in our result tables\footnote{COCO~\cite{lin2014microsoft} and ImageNet~\cite{imagenet_cvpr09} are not considered as extra data according to the MOTchallenge~\cite{MOT16,MOTChallenge20}.}: \dc{ch} for {CrowdHuman}~\cite{shao2018crowdhuman}, \dc{pt} for {PathTrack}~\cite{8237302}, \dc{re1} for the combination of {Market1501}~\cite{Zheng_2015_ICCV}, {CUHK01} and {CUHK03}~\cite{li2014deepreid} person re-identification datasets, \dc{re2} replaces {CUHK01}~\cite{li2014deepreid} with DukeMTMC~\cite{ristani2016performance}, \dc{5d1} for the use of five extra datasets (ETH~\cite{eth_biwi_00534}, {Caltech Pedestrian}~\cite{Dollar2012PAMI,dollarCVPR09peds}, {CityPersons}~\cite{zhang2017citypersons}, {CUHK-SYS~\cite{xiao2016end}}, and {PRW}~\cite{zheng2017person}), \dc{5d1+CH} is the same as \dc{5d1} plus CroudHuman.  (\dc{5d1+CH}) uses the tracking/detection results of FairMOT~\cite{zhang2020fairmot} trained within the \dc{5d1+CH} setting, and \dc{no} stands for using no extra dataset.
\PAR{Metrics.} Standard MOT metrics such as MOTA (Multiple Object Tracking Accuracy) and MOTP (Multiple Object Tracking Precision)~\cite{bernardin2008evaluating} are used: MOTA is mostly used since it reflects the average tracking performance including the number of FP (False positives, predicted bounding boxes not enclosing any object), FN (False negatives, missing ground-truth objects) and IDS~\cite{li2009learning} (Identities of predicted trajectories switch through time). MOTP evaluates the quality of bounding boxes from successfully tracked objects. Moreover, we also evaluate IDF1~\cite{ristani2016performance} (the ratio of correctly identified detections over the average number of ground-truth objects and predicted tracks), MT (the ratio of ground-truth trajectories that are covered by a track hypothesis more than 80\% of their life span), and ML (less than 20\% of their life span).
\subsection{Testset Results and Discussion}\label{sec:sota}
\PAR{MOT17.} Tab.~\ref{tab:mot17merged} presents the results obtained in the MOT17 testset. The first global remark is that most state-of-the-art methods do not evaluate under both public and private detections, and under different extra-training data settings, while we do. Secondly, \method~sets new state-of-the-art performance compared to other methods, in terms of MOTA, under \dc{CH} and no-extra training data conditions, both for public and private detections. Precisely, the increase of MOTA w.r.t. \addnote[R3Q5-2]{1}{the state-of-the-art} methods is of $8.5\%$ and $4.8\%$ (both including unpublished methods by now) for the public detection setting under \dc{CH} and no-extra training data, and of $1.7\%$ and $4.0\%$ for the private detection setting, respectively. The superiority of \method\ is remarkable in most of the metrics. We can also observe that TransCenter trained with no extra-training data outperforms, not only the methods trained with no extra data but also some methods trained with one extra dataset. Similarly, TransCenter trained on \dc{ch} performs better than seven methods trained with five or more extra datasets in the private setting, comparable to the best result in \dc{5d1+ch} (-0.3\% MOTA), showing that \method\ is less data-hungry. Moreover, trained with \dc{5d1+ch}, the performance is further improved while running at around 11 fps. Overall, these results confirm our hypothesis that \method\ with dense detection representations and sparse tracking representations produced by global relevant queries in transformers is a better choice.
\PAR{MOT20.} Tab.~\ref{tab:mot20merge} reports the results obtained in MOT20  testset. \emph{In all settings}, similar to the case in MOT17, \method\ \emph{leads the competition by a large margin compared to all the other methods}. Concretely, \method\ outperforms current methods by +19.2\%/+8.4\% in MOTA with the public/private setting trained with \dc{ch} and +11.1\%/18.8\% without extra data. From the results, another remarkable achievement of \method\ is the significant decrease of FN while keeping a relatively low FP number. This indicates that the dense representation of the detection queries can help effectively detect objects sufficiently and accurately. As for tracking, \method\ maintains low IDS numbers in MOT20 running at around 8 fps in such crowded scenes, thanks to our careful choices of QLN and the \method\ Decoder. Very importantly, to the best of our knowledge, our study is the first to report the results of \emph{all settings} on MOT20, demonstrating the tracking capacity of \method\ even in a densely crowded scenario. The outstanding results of \method\ in MOT20 further show the effectiveness of our design.

\input{tab/kitti} 
\PAR{KITTI.}Additionally, we show the results of \method\ evaluated on the KITTI dataset. \method\ significantly outperforms CenterTrack~\cite{zhou2020tracking} in pedestrian tracking (+5.3\% MOTA) while keeping a close performance in car tracking. However, the KITTI dataset is constructed in an autonomous driving scenario with only up to 15 cars and 30 pedestrians per image but some of the sequences contain no pedestrians. The sparse object locations cannot fully show the capacity of \method~to detect and track densely crowded objects.

\subsection{Efficiency-Accuracy Tradeoff Discussion} \label{sec:speed_acc_tradeoff}

\input{tab/speed_accuracy}
To have a direct idea of the better design of \method, we discuss in detail the efficiency-accuracy tradeoff comparing \method, \methodlite, and~\method-Dual to the transformer-based concurrent works -- TransTrack~\cite{sun2020transtrack} and TrackFormer~\cite{meinhardt2021trackformer}. \addnote[R4Q2]{2}{To have a fairer comparison to these methods using the (deformable) DETR encoder (D.~DETR), i.e.~ResNet-50 with a deformable transformer encoder, we leverage TransCenter-DETR, having the same encoder as theirs while keeping the rest of its structure unchanged.} Moreover, related to our method, we show superior performance compared to the center-based MOT methods -- CenterTrack~\cite{zhou2020tracking} and FairMOT~\cite{zhang2020fairmot}. The comparisons take into account the number of model parameters, the model memory footprint during inference, the inference speed (frame per second or FPS), and the MOTA performance as shown in Tab.~\ref{tab:speed_accuracy}, completed with Tab.~\ref{tab:mot17merged} and Tab.~\ref{tab:mot20merge}. Additionally, we compare the center heatmap/query responses of the aforementioned methods in Sec.~B.1 of the Supplementary Material, showing that \method\ produces sufficient and accurate outputs.

\addnote[R1Q5]{2}{Very recently, works like ByteTrack~\cite{zhang2021bytetrack} and MO3TR-PIQ~\cite{zhu2021looking} take advantage of the recent off-the-shelf object detector YOLOX~\cite{ge2021yolox} trained with extensive data-augmentation tricks such as MixUp~\cite{zhang2019bag}, which significantly improves the MOT performance. However, we do not think that the MOT performance improvement from the object detector could be considered as their contribution to the MOT community. Even though, for the scientific interest, we use the results of the YOLOX from ByteTrack~\cite{zhang2021bytetrack} as matching and birth candidates to show that \method\ can have a similar or even better result compared to ByteTrack and MO3TR-PIQ~\cite{zhu2021looking}, denoted as \method-YOLOX.}
\footnotetext[7]{MO3TR-PIQ~\cite{zhu2021looking} is not yet published and does not provide source code nor sufficient information to evaluate the IM, \#params, and the FPS.}

All results are evaluated on both MOT17 and MOT20 testsets in the private detection setting except the comparison with ByteTrack in the public detection setting (see Pub. Det. in Tab.~\ref{tab:speed_accuracy}). All the models are pretrained on CrowdHuman~\cite{shao2018crowdhuman} except for FairMOT~\cite{zhang2020fairmot} trained also on \dc{5d1+ch} datasets, and ByteTrack additionally trained on ETH~\cite{eth_biwi_00534} and CityPersons~\cite{zhang2017citypersons}. The default input image size for CenterTrack~\cite{zhou2020tracking} is $544\times960$, $608\times1088$ for FairMOT~\cite{zhang2020fairmot} and~\methodlite, $640\times1088$ for \method\, \method-Dual, \method-DETR and~\method-YOLOX. TrackFormer~\cite{meinhardt2021trackformer} and  TransTrack~\cite{sun2020transtrack} use varying input sizes with short size of $800$; ByteTrack~\cite{zhang2021bytetrack} uses an input size of $800\times1440$.
\PAR{Compared to Transformer-Based MOT.} With the respect to our concurrent works, we compare~\method\ to them both in accuracy and inference speed. Unfortunately, Trackformer~\cite{meinhardt2021trackformer} by far only shows results on MOT17 and TransTrack~\cite{sun2020transtrack} does not show results in all settings. One important remark is that \method\ systematically outperforms TransTrack and TrackFormer in both accuracy (MOTA) and speed (FPS) with a smaller model size (\# of model parameters) and less inference memory consumption \emph{in all settings}. Precisely, using the same training data,~\method\ exhibits better performance compared TransTrack by +1.7\% MOTA (by +11.2\% v.s.~TrackFormer) in MOT17 and significantly by +8.4\% in MOT20.  \addnote[R4Q2bis]{2}{The PVT~\cite{wang2022pvt} encoder can indeed bring faster inference speed and provide more meaningful image features that boost the MOT performance. However, using the same encoder as TransTrack,~\method-DETR exhibits similar inference speed and much better performance in both MOT20 (+4.3\%) and MOT17 (+0.3\%), compared to TransTrack. This indicates that the overall design of \method\ is more efficient and powerful.} Moreover, we recall that, unlike our concurrent works,~\method~leverages pixel-level dense and multi-scale detection queries to predict dense center-based heatmaps, mitigating the miss-tracking problem while keeping relatively good computational efficiency with sparse tracking queries, efficient QLN, and \method\ Decoder. \method\ demonstrates thus a significantly better efficiency-accuracy tradeoff. Finally, for different MOT applications, we provide \method-Dual, introduced to further boost the performance in crowded scenes, and \methodlite~for efficiency-critical applications.

\PAR{Compared to Center-Based MOT.} With the long-term dependencies and the dense interactions of queries in transformers, unlike pure CNN-based center MOT methods, our queries interact globally with each other and make global decisions.~\method\ exhibits much better accuracy compared to previous center-based MOT methods, CenterTrack and FairMOT~\cite{zhou2020tracking, zhang2020fairmot}. Precisely as shown in Tab.~\ref{tab:mot17merged} and Tab.~\ref{tab:speed_accuracy}, \method\ \emph{tracks more and better} compared to CenterTrack, with +8.4\% (+10.4\%) MOTA and -71,505 (-73,812) FN in MOT17 private (public) detections, \emph{using the same dense center representations and trained with the same data}. CenterTrack does not show results in MOT20 while FairMOT, similar to CenterTrack, shows good results in MOT20. It alleviates miss detections by training with much more data, leading to much fewer FN but producing much noisier detections (FP). Surprisingly, with much less training data (\dc{CH)},~\method~still~outperforms by a large margin~\cite{zhang2020fairmot} (+11.1\% MOTA shown in Tab.~\ref{tab:speed_accuracy}) even in very crowded MOT20, with \emph{cleaner detections (suppressing -74,844 FP) and better tracking associations (-2,618 IDS)} shown in Tab.~\ref{tab:mot20merge}. \addnote[R1Q5-4]{2}{Indeed, the inference speed is slower compared to CNN-based methods}, but the above comparisons have demonstrated that previous center-based MOT methods are not comparable in terms of accuracy to \method, with an acceptable fps around 11 fps for MOT17 and 8 fps for MOT20. Moreover, to adapt to applications with more strict inference constraints, \methodlite\ is introduced, keeping a better performance while having competitive inference speed compared to center-based MOT methods.

\PAR{Compared to YOLOX-Based MOT.} \addnote[R1Q5bisbis]{2}{The recent works ByteTrack~\cite{zhang2021bytetrack} and MO3TR-PIQ show significant performance gain by using the off-the-shelf object detector YOLOX~\cite{ge2021yolox}. We believe that the gain is mainly from the object detector, which is not the contribution of any aforementioned method. This claim is further confirmed by the direct comparison of \method\ and ByteTrack in public detection, i.e. without YOLOX, where \method\ outperforms the latter with 75.9\% v.s. 67.4\% MOTA in MOT17 and 72.8\% v.s.~67.0\% in MOT20 (see Tab.~\ref{tab:speed_accuracy}).} 

\addnote[R1Q5bis]{2}{For private detection, to have a fair comparison with them, we use the detection results from YOLOX (same as ByteTrack) to demonstrate that YOLOX can also significantly boost \method\ in terms of MOTA. Precisely,~\method-YOLOX shows +2.2\% MOTA compared to MO3TR-PIQ while only having a 0.5\% difference compared to ByteTrack in MOT17. We remind that ByteTrack is an offline method with offline post-processing interpolation. In very crowded scenes like MOT20, \method-YOLOX outperforms MO3TR-PIQ and ByteTrack by +5.6\% and +0.1\% in MOTA respectively. We agree that indeed there is an inference speed discrepancy with CNN-based methods compared to our transformer-based~\method. However, in terms of accuracy, \method-YOLOX shows comparable or even better results, compared to all other methods. This indicates that the design of~\method\ is beneficial with different encoders/object detectors.}

To conclude, \method\ expresses both better accuracy and efficiency compared to transformer-based methods~\cite{sun2020transtrack,meinhardt2021trackformer}; much higher accuracy numbers and competitive efficiency compared to~\cite{zhou2020tracking,zhang2020fairmot}, showing better efficiency-accuracy balance. Moreover, with the powerful YOLOX detector, \method\ can be significantly improved in terms of MOT performance, showing the potential of \method.
\input{tab/ablation_naive}
\input{tab/merged_ablation}
\subsection{Ablation Study}\label{subsec:ablation}
In this section, we first experimentally demonstrate the importance of our proposed image-related dense queries with naive DETR to MOT approaches. Then, we justify the effectiveness of our choices of QLN and \method\ Decoder (see illustrations in Fig.~\ref{fig:qln} and Fig.~\ref{fig:transformerdecoder}, respectively), considering the computational efficiency and accuracy. All results are shown in Tab.~\ref{tab:ablateMOT1720naive}. Furthermore, in  Tab.~\ref{tab:ablateMOT1720}, we ablate the impacts of removing the external Re-ID network and the NMS (Non-Maximum Suppression) during inference. Finally, we show an additional ablation of the number of decoder layers in Sec. A of the Supplementary Material. \addnote[R4Q8]{2}{For the ablation, we divide the training sets into a train-validation split, we take the first 50\% of frames as training data and test on the last 25\%. The rest 25\% of frames in the middle of the sequences are thrown to prevent over-fitting.} All the models are pre-trained on CrowdHuman~\cite{shao2018crowdhuman} and tested under the private detection setting.
\PAR{Dense Representations Are Beneficial.} We implemented a naive DETR MOT tracker with its original 100 sparse queries (from learnable embeddings initialized from noise) with the DETR-Encoder and Dual \method\ Decoder (Line 1 in Tab.~\ref{tab:ablateMOT1720naive}). To compare, the same tracker but having dense representations (43,520, i.e.~$H/4 \times W/4$) is shown in Line 2. From the results shown in Line 1-2 of Tab.~\ref{tab:ablateMOT1720naive}, we see that the limited number of queries (100, by default in~\cite{zhu2020deformable}) is problematic because it is insufficient for detecting and tracking objects, especially in very crowded scenes MOT20  (+106,467 FN, -24.8\% MOTA, compared to Line 2). This indicates that having dense representations is beneficial, especially for handling crowded scenes in MOT. Some visualization examples are shown in Sec.~\ref{subsec:qvcs}.%
\PAR{Naive Dense, Noisy Dense~v.s.~Image-Related Dense.} One naive way to alleviate the insufficient queries is to greatly increase the number of queries like in Line 2 of Tab.~\ref{tab:ablateMOT1720naive}: we drastically increase the number of queries from 100 to 43,520 (i.e.~$H/4 \times W/4$), same as the dense output of~TransCenter. Meanwhile, we compare this naive dense queries implementation to a similar one in Line 3 but with image-related dense queries like in \method. Concretely, we obtain the dense queries from QLN$_{DQ}$  (as described in Fig.~\ref{fig:qln_b_old_trctr_mt_dq}) with the memories output from DETR-Encoder. We note that the image-related implementation~has indeed 14,450 queries (i.e.~the sum of the 1/8, 1/16, 1/32, and 1/64 of the image size), and the up-scale and merge operation (see Fig.\ref{fig:dla})
\footnotetext[8]{100 noise-initialized learnable queries.} \footnotetext[9]{43,520 noise-initialized learnable queries.}\footnotetext[10]{Like other MOT methods, we observe that clipping the box size within the image size for tracking results in MOT20 improves slightly the MOTA performance. To have a fair comparison, all the results in MOT20 are updated with this technique.}forms a dense output having the same number of pixels of 43,520. Therefore, we ensure that the supervisions for the losses are equivalent\footnote{The Gaussian supervision in~TransCenter~for negative examples has values very close to 0, thus similar to the classification loss in Line 2.} for both Line 2 and 3.

The main difference between Line 2 and 3 is the queries: the proposed multi-scale dense detection queries are related to the input image where one query represents one pixel. The benefit of image-related pixel-level queries is well-discussed in Sec.~\ref{sec:methodology}. From the experimental aspect in Tab.~\ref{tab:ablateMOT1720naive}, for the noise-initialized queries in Line 2, despite the manual one-to-one matching during training, increasing the queries naively and drastically tends to predict noisier detections causing much higher FP (compared to Line 3, +2,308 and +37,850 in MOT17 and MOT20, respectively) and thus much under-performed tracking results (-12.6\% and -12.1\% MOTA for MOT17 and MOT20, respectively). Although more sophisticated implementations using sparse noise-initialized queries and Hungarian matching loss like in~\cite{sun2020transtrack} and~\cite{meinhardt2021trackformer} exhibit improved results, \method\ shows both better accuracy and efficiency, as discussed in Sec.~\ref{sec:speed_acc_tradeoff}.

\addnote[R2Q2]{2}{Furthermore, we combine \method(-Dual) with QLN$_{SE-}$ (see Fig.~\ref{fig:qln_a_proposed} in red) where the detection queries are of image size as in QLN$_{S-}$ but noise-initialized, denoted as \textit{Noise Dense}. We note that since they are noise-initialized (i.e.~unrelated to the image), it makes no sense to discard the detection attention -- DDCA (see Sec.\ref{subsec:dualdecoders}) from \method\ Decoder. The \textit{Noisy Dense} detection queries thus need to correlate with $\mathbf{M_{t}}$ to extract information from the image $t$ in \method\ Decoder and cannot speed up by removing DDCA like in \method. The result with such design in Line 18 is compared with \textit{\method-Dual} in line 17 having the same structure but with the chosen QLN$_{S-}$. We observe that the \textit{Noise Dense} queries produce a decent result thanks to the rest of \method\ designs like image-related sparse queries, \method\ Decoder, and the use of PVT encoder but they tend to have more FP caused by the noise initialization for detection queries, leading to the overall worse performance compared to \method-Dual in MOTA, IDF1, and IDS.}
\PAR{QLN Inputs.}A QLN generates tracking queries $\mathbf{TQ}$ and memories $\mathbf{TM}$; detection queries $\mathbf{DQ}$ and memories $\mathbf{DM}$. Based on the nature of tracking, we argue that $\mathbf{TQ}$ and $\mathbf{TM}$ should be obtained from information at different time steps since tracking means associating object positions in the adjacent frames. This creates two variants of QLN, namely QLN$_{M_{t}}$ with $\mathbf{TQ}$ from $\mathbf{M_{t}}$ and $\mathbf{TM}$ are from $\mathbf{M_{t-1}}$ (Line 4) and inversely, QLN$_{D-}$, where $\mathbf{TQ}$ are from $\mathbf{M_{t-1}}$ and $\mathbf{TM}$ are from $\mathbf{M_{t}}$ (Line 5). From their results, we do not observe a significant difference in terms of performance, indicating both implementations are feasible. Further, we experimentally compare QLN$_{M_{t}}$ (Line 4) and QLN$_{DQ}$ (Line 6), where the only difference is the number of FFN for outputting $\mathbf{TQ}$. With an extra FFN for $\mathbf{TQ}$, the performance is slightly improved (+2.4\% for MOT17 and +1.1\% for MOT20).
\PAR{Efficient~\method\ Decoder.} As we discuss in Sec.~\ref{subsec:dualdecoders}, the design of~\method\ Decoder can have an important impact on the computational efficiency and accuracy with different variants (Line 11-16). Precisely, comparing Line 11 and 12, we observe indeed that, with dual-decoder handling cross-attention for both detection and tracking, the performance is superior (+0.8\% MOTA for MOT17 and +1.0\% for MOT20) but the inference is slowed down by around 50\%. Balancing the efficiency and accuracy, we argue that \method\ (Line 12) is a better choice. Moreover, by removing the TQSA module (Line 13, 14), we obtain a slight inference speed up (+0.9 fps for MOT17 and MOT20) but at the cost of accuracy (-4.4\% MOTA in MOT17 and -0.6\% in MOT20). Finally, we also study the effect of sparse and dense tracking (Line 15-16), surprisingly, we find that using sparse tracking can help better associate object positions between frames  (-20 IDS in MOT17 and -131 IDS in MOT20) and in a more efficient way (+3.4 fps in MOT17 and +2.6 in MOT20).
\PAR{Efficient PVT-Encoder.} Passing from DETR-Encoder (Line 7) to PVT-Encoder (Line 8) helps get rid of the ResNet-50 feature extractor, which partially speeds up the inference from 2.1 fps to 8.1 in MOT17, and 1.2 to 6.1 fps in MOT20. Moreover, the PVT-Encoder exhibits better results which may be due to the lighter structure that eases the training (+4.2\% MOTA in MOT17 and +3.8\% in MOT20). Similarly, with PVT-Lite, we can speed up +7 fps for MOT17 and +3.5 fps for MOT20, comparing Line 9 and 10 while keeping a competitive performance.

\PAR{External Inference Overheads.} MOT methods like~\cite{bergmann2019tracking} use an external Re-ID network to extract identity features so that we can recover the objects which are temporally suspended by the tracker through appearance similarities. The Re-ID network (paired with a light-weight optical flow estimation network LiteFlowNet~\cite{hui2018liteflownet} pre-trained on KITTI~\cite{Geiger2012CVPR}) is often a ResNet-50~\cite{bergmann2019tracking}, pre-trained on object identities in MOT17 trainset. From Tab.~\ref{tab:ablateMOT1720}, we observe that this external Re-ID does help reduce IDS, especially in crowded scenes but it slows down significantly the inference. To speed up, we replace the external Re-ID features extractor simply by sampled features from memories $\mathbf{M}_{t}$, with a feature sampler like in QLN$_{S-}$ (Fig.~\ref{fig:qln_a_proposed}) using positions from detections or tracks at $t$, which is almost costless in terms of calculation and achieves comparable results to external Re-ID.

\addnote[R3Q3-1]{1}{\noindent One of the benefits of using our image-related dense representations is the discard of the one-to-one assignment process during training. Intuitively, no NMS is needed during inference. However, recall from Eq.~\ref{eq:gaussian} that the heatmaps are Gaussian distributions centered at the object centers, a 3x3 max-pooling operation is somehow needed to select the maximum response from each distribution (i.e.~object center). In practice, an NMS is only performed as in~\cite{bergmann2019tracking} within tracked objects. However, from the results shown in Tab.~\ref{tab:ablateMOT1720}, the NMS operation between tracks has little impact on the accuracy but imports important overheads. For this reason, such NMS is discarded from \method, \methodlite, and \method-Dual.}

In summary, \method\ can efficiently perform MOT with sparse tracking queries and dense detection queries operating on the proposed QLN and \method\ Decoder structures, leveraging the PVT-Encoder. The accuracy is further enhanced with the finer multi-scale features from the PVT-Encoder, the sparsity of the tracking queries as well as the chosen designs of the QLN and~\method\ Decoder. Therefore, \method\ shows a better efficiency-accuracy tradeoff compared to naive approaches and existing works.

\begin{figure*}[t!]
\centering
\begin{subfigure}{0.38\textwidth}
  \includegraphics[width=\textwidth]{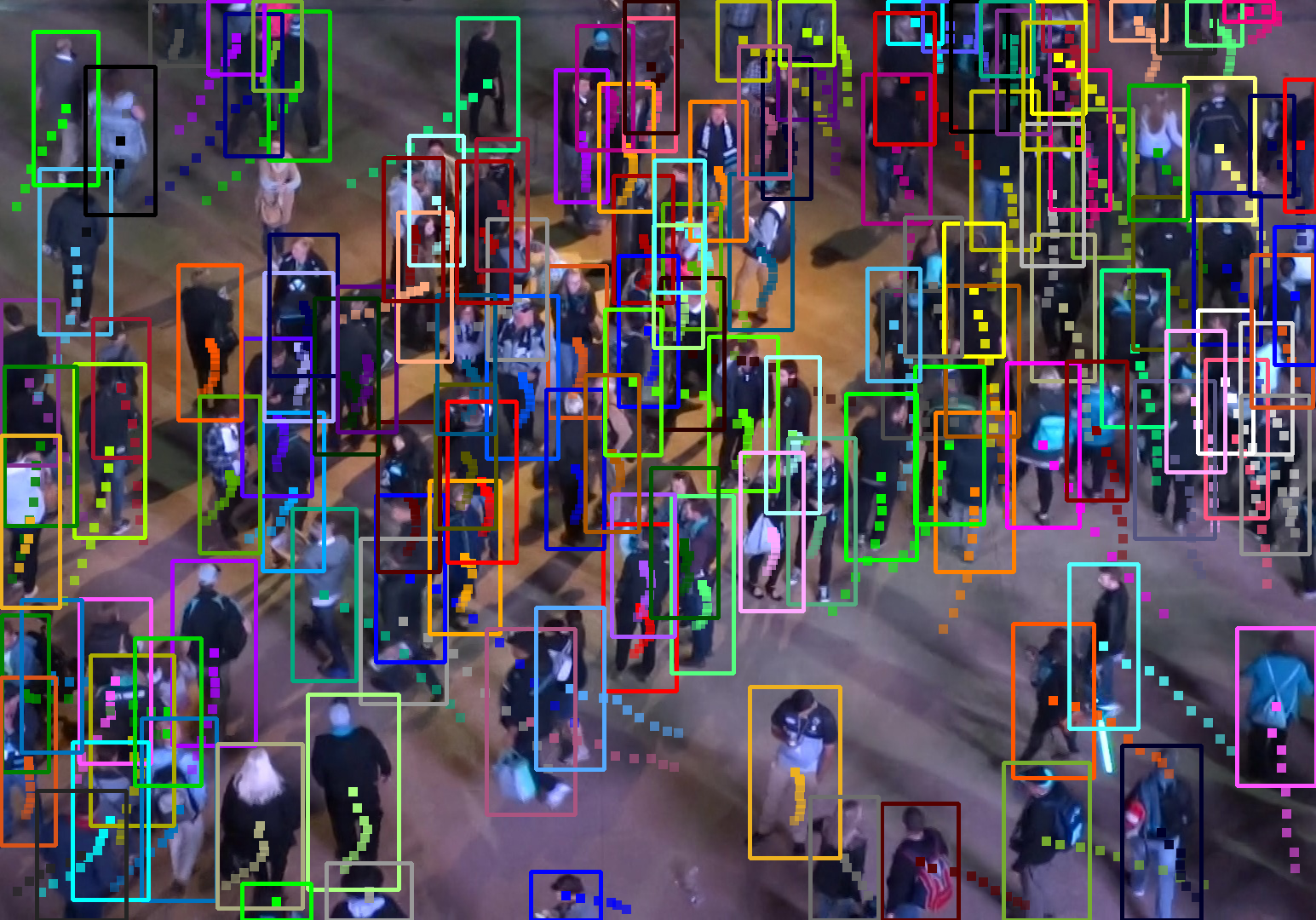}
  \caption{}
  \label{fig:MOT20-04-1-supp}
\end{subfigure}
\centering
\begin{subfigure}{0.47\textwidth}
  \includegraphics[width=\textwidth]{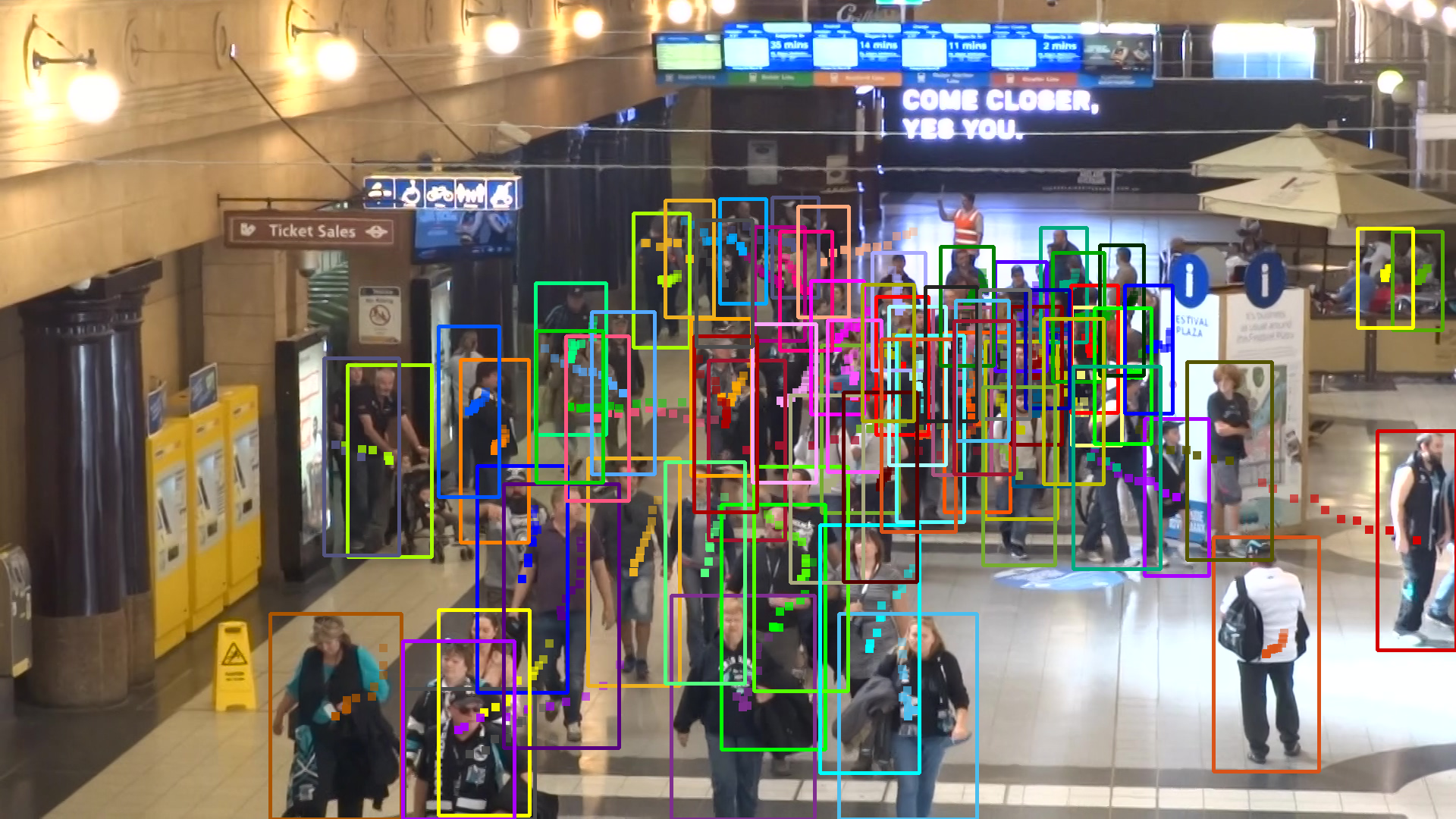}
  \caption{}
  \label{fig:MOT20-07-1-supp}
\end{subfigure}\\
\centering
\begin{subfigure}{0.86\textwidth}
  \includegraphics[width=\textwidth]{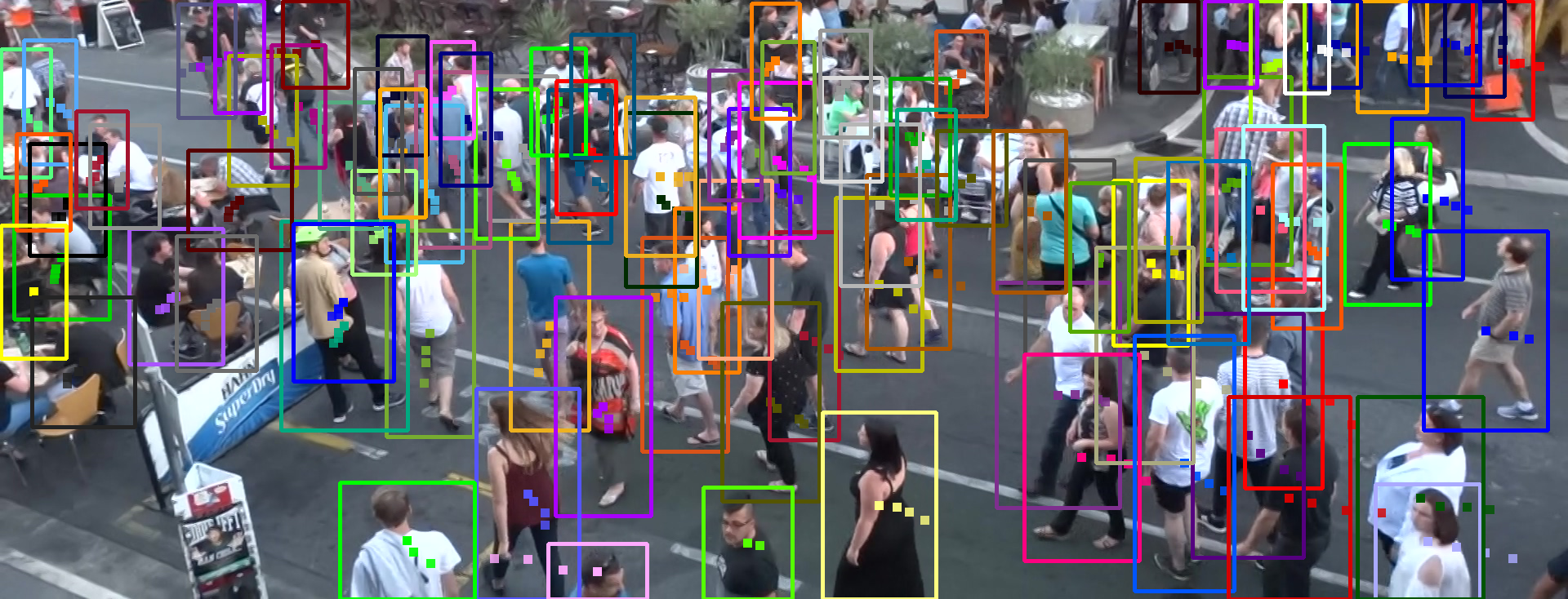}
  \caption{}
  \label{fig:MOT20-06-1-supp}
\end{subfigure}
\centering
\caption{\method\ tracking trajectories visualization of some very crowded scenes in MOT20 under the private detection setting.}\label{fig:quali-supp}
\end{figure*}
\subsection{Qualitative Visualizations in Crowded Scenes}\label{subsec:qvcs}
We report in Fig.~\ref{fig:quali-supp} the qualitative results from some crowded MOT20 sequences, to demonstrate the detection and tracking abilities of~\method\~in the context of crowded scenes. Concretely, we show in Fig.~\ref{fig:quali-supp} the predicted center trajectories and the corresponding object sizes. Fig.~\ref{fig:MOT20-04-1-supp} is extracted from MOT20-04,  Fig.~\ref{fig:MOT20-07-1-supp} from MOT20-07 and Fig.~\ref{fig:MOT20-06-1-supp} from MOT20-06. We observe that \method~manages to keep high recall, even in the context of drastic mutual occlusions, and reliably associate detections across time. To summarize,~\method~exhibits outstanding results on both MOT17  and MOT20  datasets for both public and private detections, and for both with or without extra training data, which indicates that multiple-object center tracking using transformers equipped with dense image-related queries is a promising research direction.

%% file: implementation_details.tex
\subsection{Implementation Details}\label{subsec:implementdetails}
\PAR{Inference with \method.} Once the method is trained, we detect objects by selecting the maximum responses from the output center heatmap $\mathbf{C}_t$. Since the datasets are annotated with bounding boxes, we need to convert our estimates into this representation. In detail, we apply (after max pooling) a threshold $\tau$ (e.g.~0.3 for MOT17 and 0.4 for MOT20 in \method) to the center heatmap, thus producing a list of center positions $\{\mathbf{c}_{t,k}\}_{k=1}^{K_t}$. We extract the object size $\mathbf{s}_{t,k}$ associated to each position $\mathbf{c}_{t,k}$ in $\mathbf{S}_t$. The set of detections produced by \method\ is denoted as $\mathbf{D}_t=\{\mathbf{c}_{t,k},\mathbf{s}_{t,k}\}_{k=1}^{K_t}$. In parallel, for associating objects through frames (tracking), given the position of an object $k$ at $t-1$, we can estimate the object position in the current frame by extracting the corresponding displacement estimate $\mathbf{t}_{t,k}$ from $\mathbf{T}_t$. Therefore, we can construct a set of \textit{tracked positions} $\tilde{\mathbf{P}}_{t}=\{\mathbf{c}_{t-1,k}+\mathbf{t}_{t,k},\mathbf{s}_{t,k}\}_{k=1}^{\tilde{K_t}}$. Finally, we use the Hungarian algorithm~\cite{Kuhn55thehungarian} to match the tracked positions -- $\tilde{\mathbf{P}}_{t}$ and the detection at $t$ -- $\mathbf{D}_{t}$. The matched \addnote[R2Q4-2]{2}{detections} are used to update the tracked object positions at $t$. The birth and death processes are naturally integrated in \method: detections not associated with any tracked object give birth to new tracks, while unmatched tracks are put to sleep for at most $T=60$ frames before being discarded. An external Re-ID network is often used in MOT methods~\cite{bergmann2019tracking} to recover tracks in sleep, which is proven unnecessary in our experiment in Sec.~\ref{subsec:ablation}. We also assess inference speed in fps in the testset results either obtained from~\cite{zhang2021bytetrack} or tested under the same GPU setting.
\PAR{Network and Training Parameters.} The input images are resized to $640\times 1088$ with padding in \method\ and \method-Dual while it is set to $608\times 1088$ in \methodlite. In \method, the PVT-Encoder has $[3, 4, 6, 3]$ layers ($[2, 2, 2, 2]$ in PVT-Lite) for each image feature scale and the corresponding hidden dimension $h=[64, 128, 320, 512]$ ($h=[32, 64, 160, 256]$ in PVT-Lite). $h=256$ for the \method\ Decoder with eight attention heads and six layers (four layers in \methodlite). All \method\ models are trained with loss weights $\lambda_{S}=0.1$, $\lambda_{R}=0.5$ and $\lambda_{T}=1.0$ by the AdamW optimizer~\cite{loshchilov2017decoupled} with learning rate $2\textrm{e}{-}4$. The training converges at around 50 epochs, applying a learning rate decay of $1/10$ at the 40th epoch. The entire network is pre-trained on the pedestrian class of COCO~\cite{lin2014microsoft} and then fine-tuned on the respective MOT dataset~\cite{MOT16, MOTChallenge20}.
We also present the results finetuned with extra data like CrowdHuman dataset~\cite{shao2018crowdhuman} (see Sec.~\ref{sec:sota} for details).

%% file: tab/merge_results_SOTA_MOT17.tex
\newcommand{\gb}{\cellcolor{green!12}}
\newcommand{\ob}{\cellcolor{orange!12}}
\newcommand{\rb}{\cellcolor{red!12}}
\newcommand{\bb}{\cellcolor{black!12}}

\begin{table*}[ht]
    \center
    \caption{Results on MOT17 testset: the left and right halves of the table correspond to public and private detections respectively. The cell background color encodes the amount of extra-training data: green for none, orange for one extra dataset, red for (more than) five extra datasets.  Methods with * are not associated to a publication. The best result within the same training conditions (background color) is \underline{underlined}. The best result among published methods is in \textbf{bold}. Best seen in color.} \label{tab:mot17merged}
    \tabcolsep=0.11cm
    \resizebox{\linewidth}{!}{
    
            \begin{tabular}{ l | c c c c c c c c c c|| c c c c c c c c c c}
            \toprule
            & \multicolumn{10}{c||}{Public Detections} & \multicolumn{10}{c}{Private Detections}\\\midrule
              Method & Data &  MOTA $\uparrow$ & MOTP $\uparrow$ & IDF1 $\uparrow$ & MT $\uparrow$ & ML $\downarrow$ & FP $\downarrow$ & FN $\downarrow$ & IDS $ \downarrow$ & FPS$ \uparrow$ & Data &  MOTA $\uparrow$ & MOTP $\uparrow$ & IDF1 $\uparrow$ & MT $\uparrow$ & ML $\downarrow$ & FP $\downarrow$ & FN $\downarrow$ & IDS $\downarrow$ & FPS$ \uparrow$\\ [0.5ex] 
             \midrule
                MOTDT17 \cite{chen2018real} & \ob\dc{re1}& \ob50.9 & \ob76.6 & \ob52.7 & \ob17.5 & \ob35.7 & \ob24,069 & \ob250,768 & \ob2,474  
                 & \ob \underline{\textbf{18.3}}& \bb & \bb & \bb &  \bb & \bb & \bb & \bb & \bb & \bb  & \bb \\
                 
                *UnsupTrack \cite{karthik2020simple} &  \ob\dc{pt}& \ob61.7  & \ob78.3 & \ob58.1 & \ob27.2 & \ob32.4 & \ob 16,872 & \ob197,632 & \ob 1,864 
                 & \ob $<$17.5  & \bb & \bb & \bb &  \bb & \bb & \bb & \bb & \bb & \bb  & \bb \\ %
                 
                GMT\_CT \cite{he2021learnable}& \ob\dc{re2}& \ob 61.5 & \bb & \ob {\textbf{66.9}} & \ob 26.3 & \ob 32.1 & \ob \underline{\textbf{14,059}} & \ob 200,655 & \ob \textbf{2,415}  
                 & \bb& \bb & \bb & \bb &  \bb & \bb & \bb & \bb & \bb & \bb  & \bb\\%cvpr2021 

                TrackFormer \cite{meinhardt2021trackformer} &  \ob\dc{ch} & \ob62.5 & \bb & \ob60.7 & \ob 29.8 & \ob{26.9} & \ob14,966 & \ob206,619 & \ob\underline{1,189}  & 
                 \ob 6.8 & \bb & \bb & \bb &  \bb & \bb& \ob \bb & \bb  & \ob \bb& \bb & \bb \\ %
                
                SiamMOT \cite{shuai2021siammot}& \ob\dc{ch} & \ob 65.9 & \bb & \ob 63.5 & \ob 34.6 & \ob 23.9 & \ob 18,098 & \ob 170,955 & \ob 3,040 
                  & \ob 12.8 & \bb & \bb & \bb &  \bb & \bb & \bb & \bb & \bb & \bb  & \bb \\%cvpr2021
                 
                *MOTR \cite{zeng2021motr}& \ob\dc{ch} & \ob 67.4 & \bb & \ob \underline{67.0} & \ob 34.6 & \ob 24.5 & \ob 32,355 & \ob 149,400 & \ob 1,992 
                  & \ob 7.5 & \bb & \bb & \bb &  \bb & \bb & \bb & \bb & \bb & \bb  & \bb \\%cvpr2021
                  
                TrackFormer \cite{meinhardt2021trackformer} &  \bb & \bb & \bb &\bb & \bb & \bb & \bb & \bb & \bb  & 
                 \bb& \ob\dc{ch} & \ob 65.0 & \bb &  \ob 63.9 & \ob 45.6 & \ob 13.8 & \ob70,443  & \ob 123,552& \ob 3,528 & \ob6.8 \\ %

                 CenterTrack \cite{zhou2020tracking} & \bb& \bb & \bb & \bb & \bb  & \bb  & \bb & \bb & \bb
                 &\bb  & \ob\dc{ch}& \ob67.8 &  \ob78.4 & \ob{{64.7}}& \ob34.6&  \ob24.6&  \ob\underline{\textbf{18,489}} & \ob160,332 & \ob\underline{\textbf{3,039}} 
                 & \ob \underline{\textbf{17.5}} \\%\cmidrule{11-19}
                
                TraDeS \cite{wu2021track} & \bb & \bb & \bb &  \bb & \bb & \bb & \bb & \bb & \bb  & \bb &  \ob\dc{ch}& \ob69.1 & \bb & \ob63.9 & \ob36.4 & \ob21.5 & \ob20,892 & \ob150,060 & \ob 3,555  
                 & \ob \underline{\textbf{17.5}}\\ %
                 
                PermaTrack \cite{tokmakov2021learning} & \bb & \bb & \bb &  \bb & \bb & \bb & \bb & \bb & \bb  & \bb& \ob\dc{ch} & \ob 73.8 & \bb & \ob 68.9 & \ob 43.8  & \ob 17.2  & \ob 28,998 & \ob 114,104  & \ob 3,699
                  & \ob 11.9 \\%ICCV2021
                  
                *TransTrack \cite{sun2020transtrack} & \bb & \bb & \bb &  \bb & \bb & \bb & \bb & \bb & \bb  & \bb&  \ob\dc{ch}& \ob74.5 & \ob 80.6 & \ob 63.9 & \ob46.8& \ob11.3& \ob28,323 & \ob112,137 & \ob3,663  
                 & \ob 10.0\\
                  \midrule
                
                \method\   &  \ob\dc{ch}  & \ob \underline{\textbf{75.9}}  & \ob \underline{\textbf{81.2}} & \ob 65.9  & \ob \underline{\textbf{49.8}}  & \ob \underline{\textbf{12.1}}  & \ob30,190  & \ob \underline{\textbf{100,999}} & \ob  4,626 
                & \ob 11.7 & \ob \dc{ch} &\ob \underline{\textbf{76.2}} & \ob \underline{\textbf{81.1}} &\ob \underline{\textbf{65.5}} &\ob \underline{\textbf{53.5}}  & \ob \underline{\textbf{7.9}} & 40,101\ob   & \ob \underline{\textbf{88,827}} & \ob 5,394  
                & \ob11.8  \\

                 \midrule
                 
                 GSDT \cite{Wang2021_GSDT} & \bb & \bb & \bb &  \bb & \bb & \bb & \bb & \bb & \bb & \bb &  \rb \dc{5d1} & \rb 66.2 & \rb 79.9 & \rb 68.7 & \rb 40.8 & \rb 18.3 & \rb 43,368 & \rb 144,261 & \rb 3,318 
                 & \rb 4.9\\ %
                
                SOTMOT \cite{zheng2021improving} & \rb \dc{5d1} & \rb {{62.8}} & \bb & \rb \underline{\textbf{ 67.4}} & \rb {{24.4}} & \rb {{33.0}} & \rb \underline{\textbf{6,556}} & \rb {{201,319}} & \rb \underline{\textbf{2,017}}  
                 & \rb \underline{\textbf{16.0}} & \rb \dc{5d1} & \rb 71.0 & \bb & \rb 71.9 &  \rb 42.7  & \rb 15.3 & \rb 39,537 & \rb 118,983 & \rb 5,184   & \rb 16.0\\%cvpr2021
                 
                GSDT\_V2 \cite{Wang2021_GSDT} &\bb&\bb&\bb&\bb&\bb&\bb&\bb&\bb&\bb & \bb&\rb\dc{5d1}& \rb73.2 &  \bb &  \rb66.5 & \rb41.7 & \rb17.5 & \rb \textbf{26,397} &	\rb120,666	 & \rb3,891 
                & \rb 4.9 \\ %

                CorrTracker \cite{wang2021multiple}& \bb & \bb & \bb &  \bb & \bb & \bb & \bb & \bb & \bb  & \bb & \rb \dc{5d1} & \rb \underline{\textbf{76.5}} & \bb & \rb 73.6 & \rb 47.6 & \rb {{12.7}} & \rb 29,808 & \rb{ 99,510} & \rb 3,369 
                 & \rb 15.6\\ %
                 
                FairMOT \cite{zhang2020fairmot} & \bb & \bb & \bb &  \bb & \bb & \bb & \bb & \bb & \bb  & \bb& \rb \dc{5d1+CH} & \rb 73.7 & \rb 81.3 & \rb 72.3& \rb 43.2 & \rb 17.3 & \rb 27,507 & \rb 117,477 & \rb 3,303 &  \rb \underline{\textbf{25.9}}\\ %

                *RelationTrack \cite{yu2021relationtrack} & \bb & \bb & \bb &  \bb & \bb & \bb & \bb & \bb & \bb  & \bb& \rb \dc{5d1+CH} & \rb 73.8 & \rb 81.0 & \rb 74.7 & \rb 41.7& \rb 23.2 & \rb 27,999 & \rb 118,623 & \rb \underline{1,374} &  \rb 7.4\\ %
                 
                CSTrack \cite{liang2020rethinking} & \bb & \bb & \bb &  \bb & \bb & \bb & \bb & \bb & \bb  & \bb & \rb\dc{5d1+CH}& \rb74.9 &  \rb80.9 & \rb 72.6 & \rb 41.5 & \rb 17.5 & \rb \underline{23,847}	& \rb 114,303 & \rb 3,567 &\rb 15.8\\ %
                
                MLT \cite{zhang2020multiplex} & \bb & \bb & \bb &  \bb & \bb & \bb & \bb & \bb & \bb  & \bb & \rb(\dc{5d1+CH}) & \rb 75.3  & \rb\underline{\textbf{81.7}} & \rb\underline{\textbf{75.5}} & \rb 49.3 & \rb 19.5 & \rb {27,879} & \rb 109,836 & \rb\textbf{1,719}  
                 & \rb 5.9\\ %
                
                *FUFET \cite{shan2020tracklets} &\bb& \bb & \bb & \bb &\bb&\bb& \bb &\bb & \bb 
                 & \bb & \rb(\dc{5d1+CH})& \rb 76.2 &  \rb81.1 &  \rb68.0 & \rb{{51.1}} &\rb 13.6 &\rb 32,796 &	\rb{98,475} & \rb3,237 
                 & \rb 6.8 \\
                
                  \midrule
                \method\   &  \rb\dc{5d1+CH} & \rb \underline{\textbf{76.0}} & \rb \underline{\textbf{81.4}} & \rb 65.6 & \rb \underline{\textbf{47.3}} & \rb \underline{\textbf{15.3}}  & \rb  28,369  & \rb \underline{\textbf{101,988}} & \rb 4,972   & \rb11.7 & \rb \dc{5d1+CH} & \rb 76.4 & \rb 81.2 &\rb 65.4 &\rb \underline{\textbf{51.7}} & \rb \underline{\textbf{11.6}} & \rb 37,005  & \rb \underline{\textbf{89,712}} & \rb 6,402 
                & \rb10.9
                 \\
                 \midrule
                 
                 TrctrD17 \cite{xu2020train} & \gb \dc{no} & \gb 53.7 & \gb 77.2 & \gb 53.8 & \gb 19.4 & \gb 36.6 & \gb 11,731 & \gb 247,447 & \gb 1,947  
                 & \gb $ <$2.0& \bb & \bb & \bb &  \bb & \bb & \bb & \bb & \bb & \bb  & \bb\\
                 Tracktor \cite{bergmann2019tracking} & \gb \dc{no} & \gb 53.5  & \gb 78.0 & \gb 52.3 &  \gb 19.5 & \gb 36.6 & \gb 12,201 & \gb 248,047 &  \gb 2,072  
                 & \gb $<$2.0& \bb & \bb & \bb &  \bb & \bb & \bb & \bb & \bb & \bb & \bb\\
                 
                Tracktor++ \cite{bergmann2019tracking} &\gb\dc{no}&\gb 56.3  & \gb78.8 &\gb 55.1 &\gb 21.1 & \gb 35.3 & \gb {{8,866}} & \gb 235,449 & \gb 1,987 & \gb  $<$2.0 & \bb & \bb & \bb &  \bb & \bb & \bb & \bb & \bb & \bb & \bb\\
                
                 GSM\_Tracktor \cite{ijcai2020-0074} &\gb\dc{no}& \gb56.4  & \gb77.9 & \gb57.8 & \gb22.2 &\gb 34.5 & \gb14,379 & \gb230,174 & \gb{1,485}  
                 & \gb 8.7 & \bb & \bb & \bb &  \bb & \bb & \bb & \bb & \bb & \bb & \bb\\
                 
                 TADAM \cite{guo2021online} & \gb\dc{no}& \gb 59.7 & \bb & \gb58.7 & \bb  & \bb  &  \gb \underline{\textbf{9,676}} & \gb 216,029 & \gb 1,930  
                 &\bb &\bb & \bb & \bb & \bb & \bb & \bb  &  \bb & \bb & \bb  & \bb\\%cvpr2021
                 
                CenterTrack \cite{zhou2020tracking} & \gb\dc{no}& \gb61.5 & \gb78.9 & \gb59.6 & \gb26.4  & \gb31.9  &  \gb14,076 & \gb200,672 & \gb 2,583 
                 & \gb 17.5  & \bb& \bb &  \bb & \bb& \bb&  \bb&  \bb & \bb & \bb
                 & \bb \\%\cmidrule{11-19}
                 
                 *FUFET \cite{shan2020tracklets} &\gb\dc{no}& \gb 62.0 & \bb & \gb 59.5 & \gb27.8 &\gb 31.5 & \gb15,114 &	\gb196,672 & \gb2,621  
                 & \gb 6.8 &\bb& \bb &  \bb &  \bb & \bb &\bb&\bb &	\bb & \bb
                 & \bb \\

                 ArTIST-C \cite{saleh2021probabilistic}& \gb\dc{no}& \gb62.3 & \bb & \gb 59.7 & \gb29.1 & \gb34.0 & \gb19,611 & \gb191,207 & \gb2,062  
                 & \gb $<$17.5 & \bb & \bb & \bb &  \bb & \bb & \bb & \bb & \bb & \bb 
                 &\bb \\%cvpr2021
                 
                 MAT \cite{han2020mat} & \gb\dc{no}& \gb67.1 & \gb\underline{\textbf{80.8}} & \gb\underline{\textbf{69.2}} & \gb{38.9} & \gb26.4 & \gb22,756 & \gb161,547 & \gb\underline{\textbf{1,279}}  
                 &\gb 9.0 & \bb & \bb & \bb &  \bb & \bb & \bb & \bb & \bb & \bb & \bb\\
                
                MTP \cite{kim2021discriminative} & \gb\dc{no}& \gb 51.5 & \bb & \gb54.9 & \gb20.5  & \gb35.5  &  \gb29,623 & \gb241,618 & \gb 2,563 
                & \gb 20.1 &  \gb\dc{no}& \gb 55.9 & \bb & \gb60.4 & \gb20.5 & \gb36.7  &  \gb\underline{\textbf{8,653}} & \gb238,853 & \gb \underline{\textbf{1,188}}
                & \gb 20.1 \\%cvpr2021
                
                 ChainedTracker \cite{peng2020chained} &\bb &\bb &\bb & \bb & \bb&\bb &\bb &\bb & \bb  & \bb &\gb\dc{no}&  \gb 66.6& \gb 78.2 &  \gb57.4& \gb32.2 & \gb24.2& \gb22,284 &\gb 160,491 & \gb5,529
                 & \gb 6.8\\

                 QDTrack \cite{pang2021quasi}&  \gb\dc{no}& \gb64.6 & \gb79.6 & \gb{65.1} &  \gb32.3 & \gb28.3 & \gb14,103 & \gb18,2998 & \gb2,652  
                 & \gb \underline{\textbf{20.3}} &  \gb\dc{no}& \gb68.7 & \gb79.0 & \gb\underline{\textbf{66.3}} & \gb{{40.6}}  & \gb21.9&  \gb26,589 & \gb14,6643 & \gb 3,378 
                 & \gb \underline{\textbf{20.3}} \\\midrule%
                
                \method\  & \gb\dc{no} & \gb \underline{\textbf{71.9}} & \gb 80.5 & \gb 64.1 & \gb \underline{\textbf{44.4}}  & \gb \underline{\textbf{18.6}}  & \gb 27,356  & \gb \underline{\textbf{126,860}} & \gb 4,118 
                & \gb 11.9  &  \gb\dc{no} & \gb \underline{\textbf{72.7}} & \gb \underline{\textbf{80.3}} & \gb 64.0  & \gb \underline{\textbf{48.7}}   &\gb \underline{\textbf{14.0}}  & \gb 33,807  & \gb \underline{\textbf{115,542}} & \gb  4,719
                & \gb 11.8\\

            \bottomrule
            \end{tabular}
            }

\end{table*}

%% file: tab/merge_results_SOTA_MOT20.tex
\begin{table*}[ht]
    \center
    \tabcolsep=0.11cm
    \caption{Results on MOT20 testset: the table is structured following the same principle as Tab.~\ref{tab:mot17merged}. Methods with * are not associated to a publication. The best result within the same training conditions (background color) is \underline{underlined}. The best result among published methods is in \textbf{bold}. Best seen in color.} \label{tab:mot20merge}
    \resizebox{\linewidth}{!}{
            \begin{tabular}{ l | c c c c c c c c c c || c c c c c c c c c c}
            \toprule
            & \multicolumn{10}{c||}{Public Detections} & \multicolumn{10}{c}{Private Detections}\\\midrule
              Method & Data &  MOTA $\uparrow$ & MOTP $\uparrow$ & IDF1 $\uparrow$ & MT $\uparrow$ & ML $\downarrow$ & FP $\downarrow$ & FN $\downarrow$ & IDS $\downarrow$ & FPS $\uparrow$ & Data &  MOTA $\uparrow$ & MOTP $\uparrow$ & IDF1 $\uparrow$ & MT $\uparrow$ & ML $\downarrow$ & FP $\downarrow$ & FN $\downarrow$ & IDS $\downarrow$ & FPS $\uparrow$ \\\midrule
             
            *UnsupTrack \cite{karthik2020simple} & \ob\dc{pt}& \ob53.6 & \ob{80.1} & \ob{50.6}  & \ob30.3  & \ob25.0  & \ob\underline{6,439} & \ob231,298& \ob\underline{2,178}  & \ob \underline{$<$17.5}&\bb&\bb&\bb&\bb&\bb&\bb&\bb&\bb &\bb& \bb\\
            
            *TransTrack \cite{sun2020transtrack} & \bb & \bb & \bb &  \bb & \bb & \bb & \bb & \bb  & \bb&\bb  & \ob\dc{ch}& \ob64.5 & \ob 80.0 & \ob \underline{59.2} & \ob49.1& \ob13.6& \ob \underline{28,566} & \ob151,377 & \ob3,565
            & \ob 7.2\\ \midrule

             \method   &  \ob\dc{ch} & \ob \underline{\textbf{72.8}} & \ob \underline{\textbf{81.0}}  &\ob \underline{\textbf{57.6}} &\ob \underline{\textbf{65.5}}  & \ob \underline{\textbf{12.1}} & \ob \textbf{28,026}  & \ob \underline{\textbf{110,312}} & \ob \textbf{2,621}
                & \ob \textbf{8.4} &  \ob \dc{ch} & \ob \underline{\textbf{72.9}} & \ob \underline{\textbf{81.0}} & \ob  \textbf{57.7} & \ob \underline{\textbf{66.5}} & \ob \underline{\textbf{11.8}} & \ob \textbf{28,596} & \ob \underline{\textbf{108,982}} & \ob \underline{\textbf{2,625}}  
                & \ob \underline{\textbf{8.7}}  \\

            \midrule

            CorrTracker \cite{wang2021multiple}& \bb & \bb & \bb &  \bb & \bb & \bb & \bb & \bb & \bb & \bb& \rb \dc{5d1} & \rb 65.2 & \bb & \rb 69.1 & \rb {66.4} & \rb {8.9} & \rb 79,429 & \rb \underline{\textbf{95,855}} & \rb 5,183
            & \rb 8.5\\ %

            GSDT\_V2 \cite{Wang2021_GSDT} &\bb&\bb&\bb&\bb&\bb&\bb&\bb&\bb&\bb&\bb& \rb\dc{5d1}& \rb67.1 &  \bb &  \rb67.5 & \rb53.1 & \rb13.2 & \rb{31,507} &\rb135,395	 & \rb3,230
            & \rb 0.9\\ %
            
            GSDT \cite{Wang2021_GSDT} &\bb&\bb&\bb&\bb&\bb&\bb&\bb&\bb&\bb&\bb& \rb\dc{5d1}& \rb67.1 &  \rb{79.1} &  \rb67.5 & \rb53.1 & \rb13.2 & \rb31,913 &	\rb135,409	 & \rb{{3,131}}
            & \rb 0.9\\ %
            
            SOTMOT \cite{zheng2021improving} &\bb&\bb&\bb&\bb&\bb&\bb&\bb&\bb&\bb&\bb & \rb \dc{5d1} & \rb {{68.6}} & \bb & \rb \underline{\textbf{71.4}} & \rb \underline{\textbf{64.9}} & \rb \underline{\textbf{9.7}} & \rb 57,064 & \rb 101,154 & \rb 4,209
            & \rb 8.5\\%cvpr2021

            FairMOT \cite{zhang2020fairmot}&\bb&\bb&\bb&\bb&\bb&\bb&\bb&\bb&\bb& \bb& \rb\dc{5d1+CH}& \rb61.8 & \rb78.6 &  \rb\textbf{67.3}  & \rb\underline{\textbf{68.8}}  & \rb\underline{\textbf{7.6}}  & \rb103,440& \rb \underline{\textbf{88,901}} & \rb5,243      & \rb \underline{\textbf{13.2}}\\  %
            CSTrack \cite{liang2020rethinking}&\bb&\bb&\bb&\bb&\bb&\bb&\bb&\bb&\bb&\bb&  \rb\dc{5d1+CH}& \rb66.6 &  \rb78.8 &  \rb68.6 & \rb50.4 & \rb15.5 &\rb\underline{25,404}&\rb144,358 & \rb3,196
             & \rb 4.5\\ %
             
            *RelationTrack \cite{yu2021relationtrack} & \bb & \bb & \bb &  \bb & \bb & \bb & \bb & \bb & \bb  & \bb& \rb \dc{5d1+CH} & \rb 67.2 & \rb 79.2 & \rb \underline{70.5} & \rb 62.2& \rb 8.9 & \rb 61,134 & \rb 104,597 & \rb 4,243 &  \rb 2.7\\ %
                        \midrule
            
            \method   &  \rb\dc{5d1+CH} & \rb \underline{\textbf{72.4}} &\rb \underline{\textbf{81.2}} & \rb \underline{\textbf{57.9}} &\rb \underline{\textbf{64.2}} & \rb \underline{\textbf{12.3}} & \rb \underline{\textbf{25,121}}   & \rb \underline{\textbf{115,421}} & \rb  \underline{\textbf{2,290}}
                & \rb \underline{\textbf{8.6}} &  \rb \dc{5d1+CH} & \rb \underline{\textbf{72.5}} & \rb \underline{\textbf{81.1}} & \rb 58.1  & \rb 64.7 & \rb 12.2  & \rb {\textbf{25,722}} & \rb  114,310 & \rb \underline{\textbf{2,332}}
                & \rb 8.8  \\
                 \midrule

            SORT \cite{bewley2016simple}&\gb\dc{no}& \gb42.7  & \gb78.5 & \gb45.1  & \gb16.7  & \gb26.2  &\gb27,521 &	\gb264,694	 & \gb4,470  
            \gb& \gb \underline{\textbf{ $<$27.7}} &\bb&\bb&\bb&\bb&\bb&\bb&\bb&\bb&\bb& \bb\\

            Tracktor++ \cite{bergmann2019tracking} &\gb\dc{no}& \gb52.6 & \gb\underline{\textbf{79.9}} & \gb{{52.7}}  & \gb29.4  & \gb26.7  & \gb \underline{{\textbf{6,930}}} &\gb236,680 & \gb1,648  
            & \gb 1.2 &\bb&\bb&\bb&\bb&\bb&\bb&\bb&\bb&\bb& \bb\\
            
            ArTIST-T \cite{saleh2021probabilistic}& \gb\dc{no}& \gb53.6 & \bb & \gb 51.0 & \gb31.6 & \gb28.1 & \gb7,765 & \gb230,576 & \gb \underline{\textbf{1,531}} 
            & \gb $<$1.2& \bb & \bb & \bb &  \bb & \bb & \bb & \bb & \bb & \bb & \bb\\ %
            
            GNNMatch \cite{papakis2020gcnnmatch} &\gb\dc{no}& \gb54.5& \gb79.4 & \gb49.0 & \gb32.8 & \gb25.5 & \gb9,522 & \gb223,611& \gb2,038 
            & \gb 0.1 &\bb&\bb&\bb&\bb&\bb&\bb&\bb&\bb&\bb & \bb\\
             
             TADAM \cite{guo2021online} & \gb\dc{no}& \gb 56.6 & \bb & \gb51.6 & \bb  & \bb  &  \gb39,407 & \gb 18,2520 & \gb 2,690 
             &\bb &\bb & \bb & \bb & \bb & \bb & \bb  &  \bb & \bb & \bb& \bb\\%cvpr2021

            MLT \cite{zhang2020multiplex} &\bb&\bb&\bb&\bb&\bb&\bb&\bb&\bb&\bb& \bb&\gb\dc{no}& \gb48.9 & \gb78.0 & \gb{{54.6}}  & \gb30.9  & \gb22.1  & \gb\underline{\textbf{45,660}} & \gb216,803 & \gb\underline{\textbf{2,187}}& \gb 3.7 \\\midrule
             
            \method  & \gb\dc{no} & \gb \underline{\textbf{ 67.7}}  & \gb 79.8  & \gb \underline{\textbf{58.9}}   & \gb \underline{\textbf{65.6}}  & \gb \underline{\textbf{11.3}}   & \gb 54,967   & \gb \underline{\textbf{108,376}} & \gb 3,707
                & \gb 8.4  &  \gb\dc{no} & \gb \underline{\textbf{ 67.7}}  & \gb \underline{\textbf{79.8}}   & \gb \underline{\textbf{58.7}} & \gb \underline{\textbf{66.3}} &\gb \underline{\textbf{11.1}}  & \gb  56,435  & \gb \underline{\textbf{107,163}}  & \gb 3,759 
                & \gb \underline{\textbf{8.4}} \\
            
            \bottomrule
            \end{tabular}
            }
\end{table*}

%% file: tab/kitti.tex
\begin{table}[ht]
\centering

\tabcolsep=0.11cm
    \caption{KITTI testset results in MOTA, MOTP, FP, FN, IDS and FPS. Best results are \underline{underlined}.}
    \resizebox{\linewidth}{!}{
    \begin{tabular}{c l| c c c c c c}
    \toprule
    \midrule
     &Method & MOTA $\uparrow$ & MOTP $\uparrow$ & FP $\downarrow$ & FN $\downarrow$ & IDS $\downarrow$  & FPS $\uparrow$ \\ [0.5ex] 
     \midrule
     \parbox[t]{1mm}{\multirow{7}{*}{\rotatebox[origin=c]{90}{Car}}} &   MASS~\cite{karunasekera2019multiple} & 84.6 & 85.4 & 4,145 & 786 & 353 & 100.0  \\
     &  IMMDP~\cite{xiang2015learning}       & 82.8 & 82.8 & 5,300 & 422 & 211 & 5.3  \\
     &  AB3D~\cite{weng20203d}               &83.5  & 85.2 &4,492  &1,060  & \underline{126} & \underline{214.7}  \\
     &  SMAT~\cite{gonzalez2020smat}         & 83.6 & 85.9 & 5,254 & \underline{175} & 198 &10.0 \\
     &  TrackMPNN~\cite{rangesh2021trackmpnn}     & 87.3 & 84.5 & 2,577 & 1,298 & 481 & 20.0\\
     &  CenterTrack~\cite{zhou2020tracking}   & 88.8 & 85.0 & 2,703 & 886 & 254  & 22.2\\
       \cmidrule(){2-8}
     & \method   & 87.3 & 83.8  & 3,189 &  847 & 340  & 18.5 \\
    
    \midrule
     \parbox[t]{1mm}{\multirow{4}{*}{\rotatebox[origin=c]{90}{Person}}} &  AB3D~\cite{weng20203d}  & 38.9 & 64.6 &  11,744 & 2,135   & 259 &  \underline{214.7}  \\
     &  TrackMPNN~\cite{rangesh2021trackmpnn}  & 52.1 &73.4  &7,705   &  2,758  & 626  &  20.0 \\
     &  CenterTrack~\cite{zhou2020tracking}   & 53.8  & 73,7 &  8,061 &  2,201  & 425 & 22.2   \\
       \cmidrule(){2-8}
     & \method  & 59.1 &  73.2 & 6,889 & 2,142  & 436   &  18.5\\
    \bottomrule
    \end{tabular}
    }
  \label{tab:kitti}
\end{table}

%% file: tab/speed_accuracy.tex
\begin{table}[ht]
\centering
\tabcolsep=0.11cm
    \caption{Testset comparisons on the MOT17 and MOT20 among CenterTrack~\cite{zhou2020tracking}, FairMOT~\cite{zhang2020fairmot},
    Trackformer~\cite{meinhardt2021trackformer},
    TransTrack~\cite{sun2020transtrack}, ByteTrack~\cite{zhang2021bytetrack}, MO3TR-PIQ~\cite{zhu2021looking} and our proposed models in number of model parameters (\#params), Inference Memory (IM), Frame Per Second (FPS) and MOTA. Best results are \underline{underlined}. They are grouped according to their encoders or used detections, indicated as Enc/Det. Without extra notation, private detections are used by default.}
    \resizebox{\linewidth}{!}{
    \begin{tabular}{c c c c c c  c}
    \toprule

     & Model & Enc/Det & \#params (M) $\downarrow$ & IM (MB) $\downarrow$  &  FPS $\uparrow$ & MOTA $\uparrow$\\[0.5ex]
     \midrule
   \parbox[t]{1mm}{\multirow{13}{*}{\rotatebox[origin=c]{90}{MOT-17}}}
   
    &FairMOT~\cite{zhang2020fairmot}& \multirow{2}{*}{DLA-34}& 20.3 & 892  & 25.9 & 73.7 \\
    &CenterTrack~\cite{zhou2020tracking} & &  20.0 & 892  &17.5 & 67.8  \\
    
    \cmidrule(r){2-7}
    &TrackFormer~\cite{meinhardt2021trackformer} &\multirow{3}{*}{D. DETR} &40.0 & 976  & 6.8  & 65.0 \\
    &TransTrack~\cite{sun2020transtrack} & & 47.1 &  1,002 &10.0 & 74.5 \\
    &TransCenter-DETR  & & 43.6 & 980 & 8.8 & 74.8 \\
    \cmidrule(r){2-7}
    
    &ByteTrack~\cite{zhang2021bytetrack} & \multirow{3}{*}{YOLOX} & 99.1 & 1,746  & \underline{29.6} & \underline{80.3} \\
    &MO3TR-PIQ~\cite{zhu2021looking}\protect \footnotemark & & N/A & N/A & N/A & 77.6 \\
    &TransCenter-YOLOX & & 35.1 & 938 &  11.9 & 79.8\\
    
    \cmidrule(r){2-7}
    &TransCenter-Dual  &\multirow{3}{*}{PVT} & 39.7 & 954 &5.6& 76.0 \\
    &TransCenter  & & 35.1 & 938 & 11.8& 76.2\\
    &TransCenter-Lite & & \underline{8.1} &  \underline{838}   &17.5 & 73.5\\
    
    \cmidrule(r){2-7}
    &ByteTrack  &\multirow{2}{*}{Pub. Det.} & 99.1 & 1,746 & N/A & 67.4 \\
    &TransCenter  & & 35.1 & 938 & 11.7 & 75.9\\

    \midrule \midrule
    \parbox[t]{1mm}{\multirow{13}{*}{\rotatebox[origin=c]{90}{MOT-20}}}
    
    &FairMOT~\cite{zhang2020fairmot} & \multirow{2}{*}{DLA-34}  & 20.3 &  892   &  13.2 &  61.8   \\
    &CenterTrack~\cite{zhou2020tracking} & & N/A  &N/A & N/A & N/A  \\
    
    \cmidrule(r){2-7}
    &TrackFormer~\cite{meinhardt2021trackformer} & \multirow{3}{*}{D. DETR} & N/A & N/A & N/A & N/A \\
    &TransTrack~\cite{sun2020transtrack} & & 47.1 &  1,002  & 7.2  & 64.5 \\
    &TransCenter-DETR & & 43.6 & 980 & 7.5 & 68.8\\
    
    \cmidrule(r){2-7}
    &ByteTrack~\cite{zhang2021bytetrack} &\multirow{3}{*}{YOLOX} & 99.1 & 1,746  & \underline{17.5} & 77.8 \\
    &MO3TR-PIQ~\cite{zhu2021looking}  & & N/A & N/A & N/A & 72.3\\
    &TransCenter-YOLOX & & 35.1 & 938 & 10.1 & \underline{77.9}\\
    
    \cmidrule(r){2-7}
    &TransCenter-Dual  & \multirow{3}{*}{PVT} &  39.7 & 954 &5.1& 73.5 \\
    &TransCenter  & &  35.1 & 938 &8.7 & 72.9\\
    &TransCenter-Lite & & \underline{8.1} & \underline{838}  & 11.0 & 68.0  \\
    \cmidrule(r){2-7}
    &ByteTrack  &\multirow{2}{*}{Pub. Det.} & 99.1 & 1,746 & N/A & 67.0\\
    &TransCenter  & & 35.1 & 938 & 8.4 & 72.8\\
    \bottomrule
    \end{tabular}
    }
  \label{tab:speed_accuracy}
\end{table}

%% file: tab/ablation_naive.tex
\begin{table*}[ht]
\centering
\tabcolsep=0.11cm
    \caption{Ablation study of different model structure implementations evaluated on MOT17 and MOT20 validation sets: different variants of the QLN and \method-decoders are discussed in Fig.~\ref{fig:qln} and Fig.~\ref{fig:transformerdecoder}, respectively. The module ablated in each line is marked in \colorbox{black!12}{grey}. For the queries, "D-S“ means dense detection and sparse tracking queries, with "D" refers to "dense" and "S" refers to "sparse", similarly for "S-S" and "D-D". Line 10, 12, 14, 16 are the proposed \method, line 9 is~\methodlite\ and line 11, and 17 are~\method-Dual, indicated with $*$.}
    \resizebox{\textwidth}{!}{
    \begin{tabular}{c c c c c| c c c c c c | c c c c c c c}
    \toprule
    
     & \multicolumn{3}{c}{Settings}  && & \multicolumn{4}{c}{MOT17}  & & \multicolumn{6}{c}{MOT20} \\ \midrule

      Line&Encoder & Decoder & Queries & QLN  & MOTA $\uparrow$ & IDF1 $\uparrow$ & FP $\downarrow$ & FN $\downarrow$ & IDS $\downarrow$  & FPS $\uparrow$  & MOTA $\uparrow$ & IDF1 $\uparrow$ & FP $\downarrow$ & FN $\downarrow$ & IDS $\downarrow$  & FPS $\uparrow$ \\[0.5ex] 
     \midrule
      1& DETR &Dual& \bb S-S\protect \footnotemark & \bb QLN$_{E}$  & 49.4  & 49.6 &  4,909 & 8,202 & 602 & \underline{2.5} & 42.5 & 26.8 &19,243  &   153,429  &  4,375 & \underline{1.5} \\ %

      2&DETR &Dual& \bb D-D\protect \footnotemark & \bb QLN$_{E}$ & 56.6 & 50.1  &3,510 & 7,577 & 678 & 1.7 &67.3  &32.1  & 45,547 & \underline{46,962} & 8,115 &0.9 \\ \vspace{+1mm}%
      3&DETR & Dual & \bb D-D& \bb QLN$_{DQ}$ & \underline{69.2} & \underline{71.9} & \underline{1,202}  & \underline{6,951} & \underline{203} &1.9& \underline{79.4}\protect \footnotemark & \underline{66.9} & \underline{7,697} & 53,987 & \underline{1,680}&1.2 \\ %
      \midrule  
       4& DETR&Dual &D-D& \bb QLN$_{M_t}$  & 66.8  & 71.2  & 1,074  & 7,757 &  \underline{167}&1.9 &  78.3 & 64.7  & 5517 & 59,447 & 1,832 & 1.2 \\ %

        5&DETR& Dual & D-D & \bb QLN$_{D-}$  & 65.8  & 70.3  & \underline{1,122} & 7,987  & 176 &1.9 &  78.4& 64.2 & \underline{5,340} &59,288 & 1,798& 1.1 \\ %
       6&DETR & Dual & D-D& \bb QLN$_{DQ}$ & \underline{69.2} & \underline{71.9} & 1,202  & \underline{6,951} & 203 &1.9& \underline{79.4} & \underline{66.9} & 7,697 & 53,987 & \underline{1,680}&1.2 \\ %

         \midrule
     7&\bb DETR& Single & D-D &QLN$_{DQ}$ & 68.1  & \underline{72.0} &  \underline{580} &  7,922 & \underline{141} & 2.1& 79.8 & 66.9 & \underline{6,955}  & 53,445 & 1,657& 1.2\\ %
     
     8&\bb PVT& Single & D-D& QLN$_{DQ}$ & \underline{72.3} & 71.4  & 1,116  & \underline{6,156} & 249 & \underline{8.1} & \underline{83.6}  & \underline{75.1} &  14,358 & \underline{34,782}& \underline{1,348} & \underline{6.1}\\ %
     
        \midrule
      9$^*$&\bb PVT-Lite& Single &D-S& QLN$_{S-}$  & 69.8 & 71.6  & 2,008  & 5,923  & \underline{252}&\underline{19.4}& 82.9 & 75.5  & \underline{14,857} & 36,510 & 1,181&\underline{12.4} \\ %
       
      10$^*$&\bb PVT & TQSA-Single & D-S & QLN$_{S-}$  & \underline{73.8} & \underline{74.1} & \underline{1,302}  & \underline{5,540}  & 258& 12.4&  \underline{83.6}& \underline{75.7} & 15,459 & \underline{34,054} & \underline{1,085} & 8.9\\ %
     
       \midrule
       11$^*$&PVT &\bb TQSA-Dual & D-S & QLN$_{S-}$  & \underline{74.6}  & \underline{76.5} &  \underline{892} & 5,879  & \underline{128} &5.6& \underline{84.6} & \underline{78.0} & \underline{13,415} & \underline{33,202} & \underline{921} &4.8\\ %
       
       12$^*$&PVT & \bb TQSA-Single & D-S & QLN$_{S-}$  & 73.8 & 74.1 & 1,302  & \underline{5,540}  & 258& \underline{12.4}&  83.6& 75.7 & 15,459 & 34,054 & 1,085 & \underline{8.9}\\ %
    
    \midrule
    13&PVT & \bb Single & D-S & QLN$_{S-}$ & 69.4 & 72.5  & 1,756  & 6,335 & \underline{197} & \underline{13.3} & 83.0  &74.7 &  \underline{14,584} & 36,441& 1,321 &\underline{9.8} \\ %
  14$^*$&PVT & \bb TQSA-Single & D-S& QLN$_{S-}$ & \underline{73.8}  & \underline{74.1} & \underline{1,302} & \underline{5,540} & 258 & 12.4 & \underline{83.6} & \underline{75.7} & 15,459 & \underline{34,054} & \underline{1,085}  & 8.9 \\ %
  
      \midrule
    15&PVT &  Single & \bb D-D & QLN$_{D-}$ & 71.1 & 71.7  & \underline{1,274}  & 6,274 & 278 & 9.0 & 82.5  & 75.0 &  \underline{12,686} & 40,003 & 1,216 & 6.3 \\ %
  16$^*$&PVT & TQSA-Single & \bb D-S& QLN$_{S-}$ & \underline{73.8}  & \underline{74.1} & 1,302 & \underline{5,540} & \underline{258} & \underline{12.4} & \underline{83.6} & \underline{75.7} & 15,459 & \underline{34,054} & \underline{1,085}  & \underline{8.9} \\ %
  
    \midrule
   17$^*$&PVT & TQSA-Dual & D-S & \bb QLN$_{S-}$  & \underline{74.6}  & \underline{76.5} &  \underline{892} & 5,879  & \underline{128} &\underline{5.6}& \underline{84.6} & \underline{78.0} & \underline{13,415} & 33,202 & \underline{921} &\underline{4.8}\\ %
   
   18&PVT & TQSA-Dual & D-S & \bb QLN$_{SE-}$  &  72.7 & 73.7  &  2,012  & \underline{5,213}  & 181 & 5.5 & 83.9 & 77.4 & 16,252 & \underline{32,402} & 976 & \underline{4.8}\\

    \bottomrule
    \end{tabular}
    }
  \label{tab:ablateMOT1720naive}
\end{table*}

%% file: tab/merged_ablation.tex
\begin{table*}[ht]
\centering
\tabcolsep=0.11cm
    \caption{Ablation study of external inference overheads: the results are expressed in MOTA, MOTP, FP, FN, IDS as well as FPS. They are evaluated on both MOT17 and MOT20 validation sets using \method, \methodlite\ and \method-Dual, respectively.}
    \resizebox{.8\linewidth}{!}{
    \begin{tabular}{ c| c c c c c c | c c c c c c c}
    \toprule
    
       &   &  \multicolumn{4}{c}{MOT17}  &  & &  & MOT20  &    & &   \\
    \midrule

     Setting & MOTA $\uparrow$ & IDF1 $\uparrow$ & FP $\downarrow$ & FN $\downarrow$ & IDS $\downarrow$  & FPS $\uparrow$  & MOTA $\uparrow$ & IDF1 $\uparrow$ & FP $\downarrow$ & FN $\downarrow$ & IDS $\downarrow$  & FPS $\uparrow$ \\[0.5ex] 
    \midrule

      + ex-ReID  & 73.7  & 73.5 &  \underline{1,047} & 5,800  & 286 & 4.5 & \underline{83.6} & 75.5 & 15,468 & 34,057 & \underline{1,064} & 2.9 \\
    
     + NMS & \underline{73.8}  & \underline{74.1} &  1,300 & 5,548 & 261& 10.8 &\underline{83.6} & \underline{75.8} & \underline{15,458} & \underline{34,053} & 1,084 & 5.1 \\
      \method\ & \underline{73.8}  & \underline{74.1} & 1,302 & \underline{5,540} & \underline{258} & \underline{12.4} & \underline{83.6} & 75.7& 15,459 & 34,054 & 1,085  & \underline{8.9} \\
     
     \midrule
    
     + ex-ReID  &  \underline{69.8}  & \underline{72.0}  & \underline{2,008} &  \underline{5,922} & \underline{248} & 5.2 & \underline{82.9} & 74.9  & 14,894 & \underline{36,498} & \underline{1,147} & 3.1\\
      + NMS &\underline{69.8}  & 71.6 & 2,009 & 5,923 & 253 & 15.9  &\underline{82.9} & 75.4 & 14,872 & 36,507   &  1,181 & 6.2\\
      \methodlite\ & \underline{69.8} & 71.6 & \underline{2,008} & 5,923 & 252 & \underline{19.4} & \underline{82.9} & \underline{75.5}& \underline{14,857}  & 36,510& 1,181  & \underline{12.4}\\

    \midrule

     + ex-ReID    & \underline{74.6} & 75.8  & \underline{840}  & 5,882 & 162  & 3.0  & 84.5 & 77.7 & 13,478 & 33,209& \underline{913} & 2.2 \\
    
     + NMS  & \underline{74.6}  & \underline{76.5}  & 891 & \underline{5,879} &  \underline{127}  & 5.1 & \underline{84.6} & 77.9  & \underline{13,413} & 33,203 & 921 & 3.4 \\
      \method-Dual  & \underline{74.6}  & \underline{76.5} &  892 & \underline{5,879} &  128 & \underline{5.6} & \underline{84.6} & \underline{78.0} & 13,415 & \underline{33,202} & 921 & \underline{4.8} \\
     
    \bottomrule
    \end{tabular}
    }
  \label{tab:ablateMOT1720}
\end{table*}

%% file: conclusion.tex
\section{Conclusion}
In this paper, we introduce \method, a powerful and efficient transformer-based architecture with dense image-related representations for multiple-object tracking. In  \method, we propose the use of dense pixel-level multi-scale detection queries and sparse tracking queries. They are produced with carefully-designed QLN and interacting in~\method\ Decoder with image features. Under the same training conditions, \method\ outperforms its competitors in MOT17 and by a large margin in very crowded scenes in MOT20. It even exhibits better performance than some methods trained with much more data. More importantly, \method\ maintains a reasonably high efficiency while exhibiting a state-of-the-art performance, thanks to its careful designs proven by a thorough ablation.

%% file: supple.tex
\vspace{3mm}
{\Large
\begin{center}
\textbf{Supplementary Material}    
\end{center}
}\vspace{3mm}
\begin{figure}[ht]
\centering
\begin{subfigure}{0.24\textwidth}
  \includegraphics[width=\textwidth]{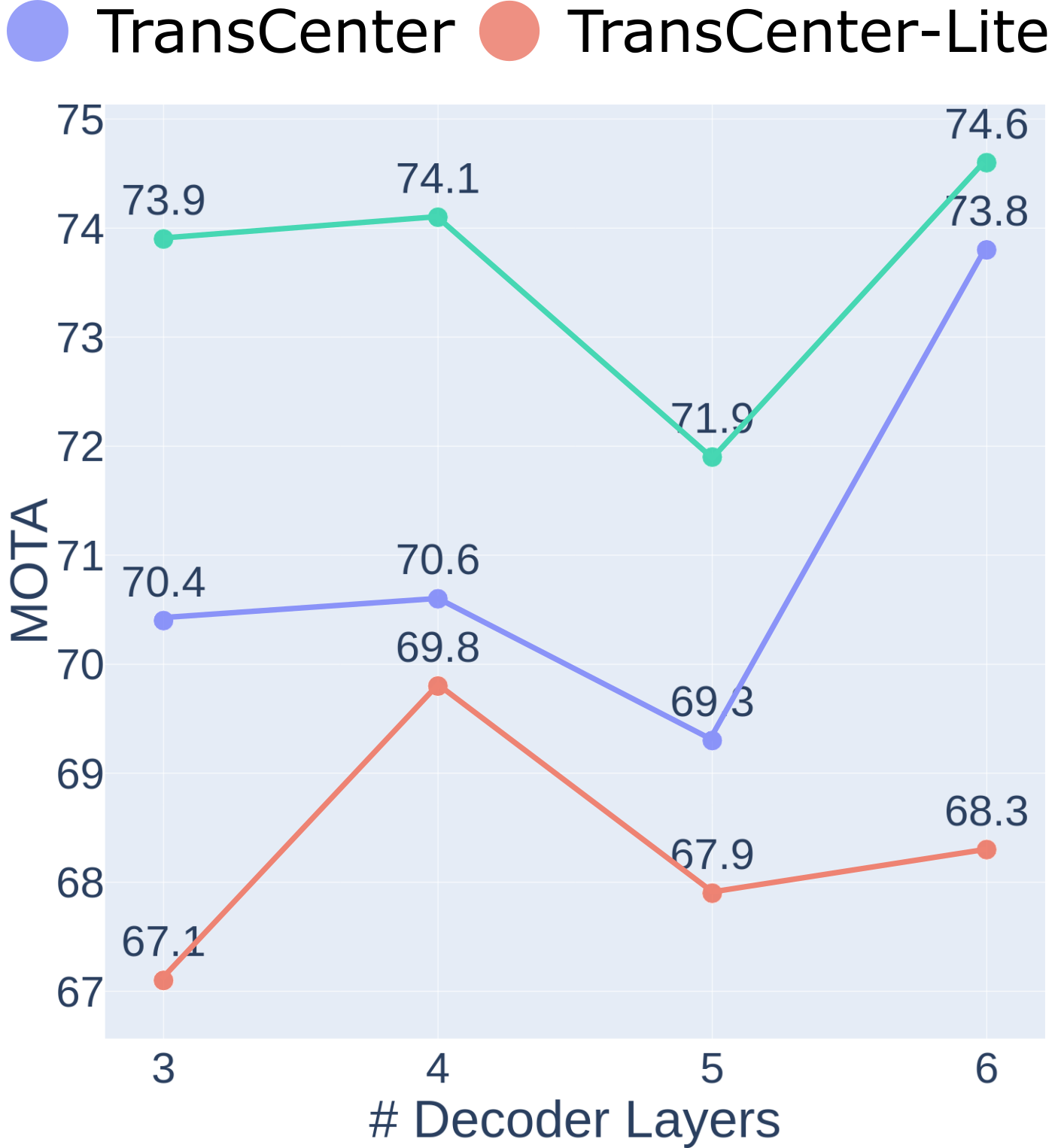}\vspace{-1.5mm}
  \caption{MOT17}
  \label{fig:ablate_num_layers_mot17}
\end{subfigure}
\begin{subfigure}{0.24\textwidth}
  \includegraphics[width=\textwidth]{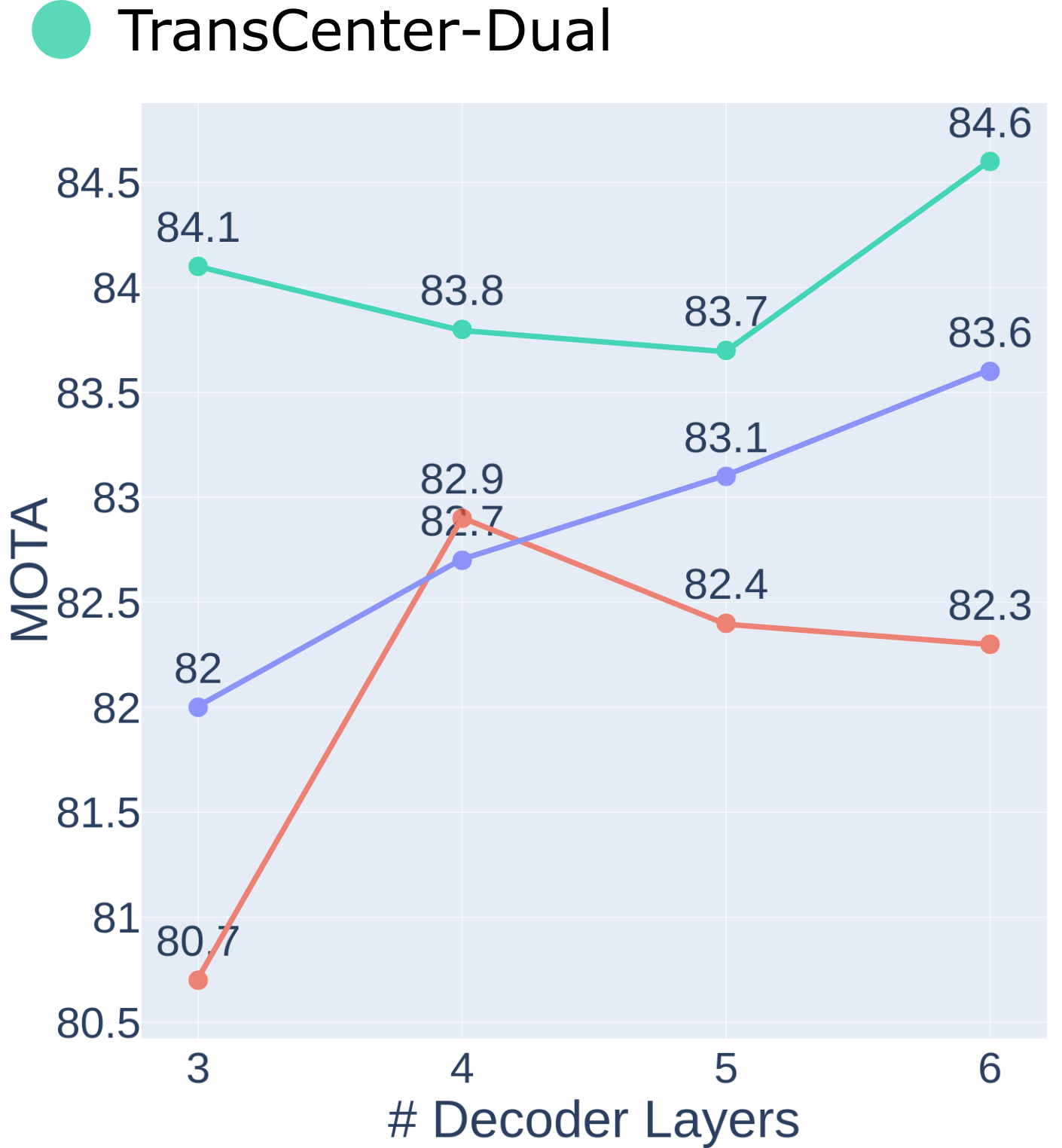}\vspace{-1.5mm}
  \caption{MOT20}
  \label{fig:ablate_num_layers_mot20}
\end{subfigure}\hspace{1mm}%
\caption{Ablation results in MOTA using different numbers of \method\ Decoder layers. The results are evaluated on both MOT17 and MOT20 validation sets with \method, \methodlite\ and \method-Dual.}\label{fig:ablate_num_layers}
\end{figure}\vspace{-5mm}
\section{Number of Decoder Layers} We ablate in Fig.~\ref{fig:ablate_num_layers} the number of decoder layers in \method, \methodlite, and~\method-Dual. Precisely, we search the best number of decoder layers within the range of $[3,6]$ For~\method, \method-Dual, we find that having six layers of decoder gets the best results in terms of MOTA both in MOT17 and MOT20 while with four layers in \methodlite.

\section{Qualitative Results and Visualizations}\label{subsec:quali}
In this section, we qualitatively visualize the center heatmap responses (Sec.~\ref{subsec:chr}), the detection queries (Sec.~\ref{subsec:dqv}) and the tracking queries (Sec.~\ref{subsec:tqv}) of~\method. 
\begin{figure}[t]
\centering
\begin{subfigure}{0.24\textwidth}
  \includegraphics[width=\textwidth,height=2.2cm]{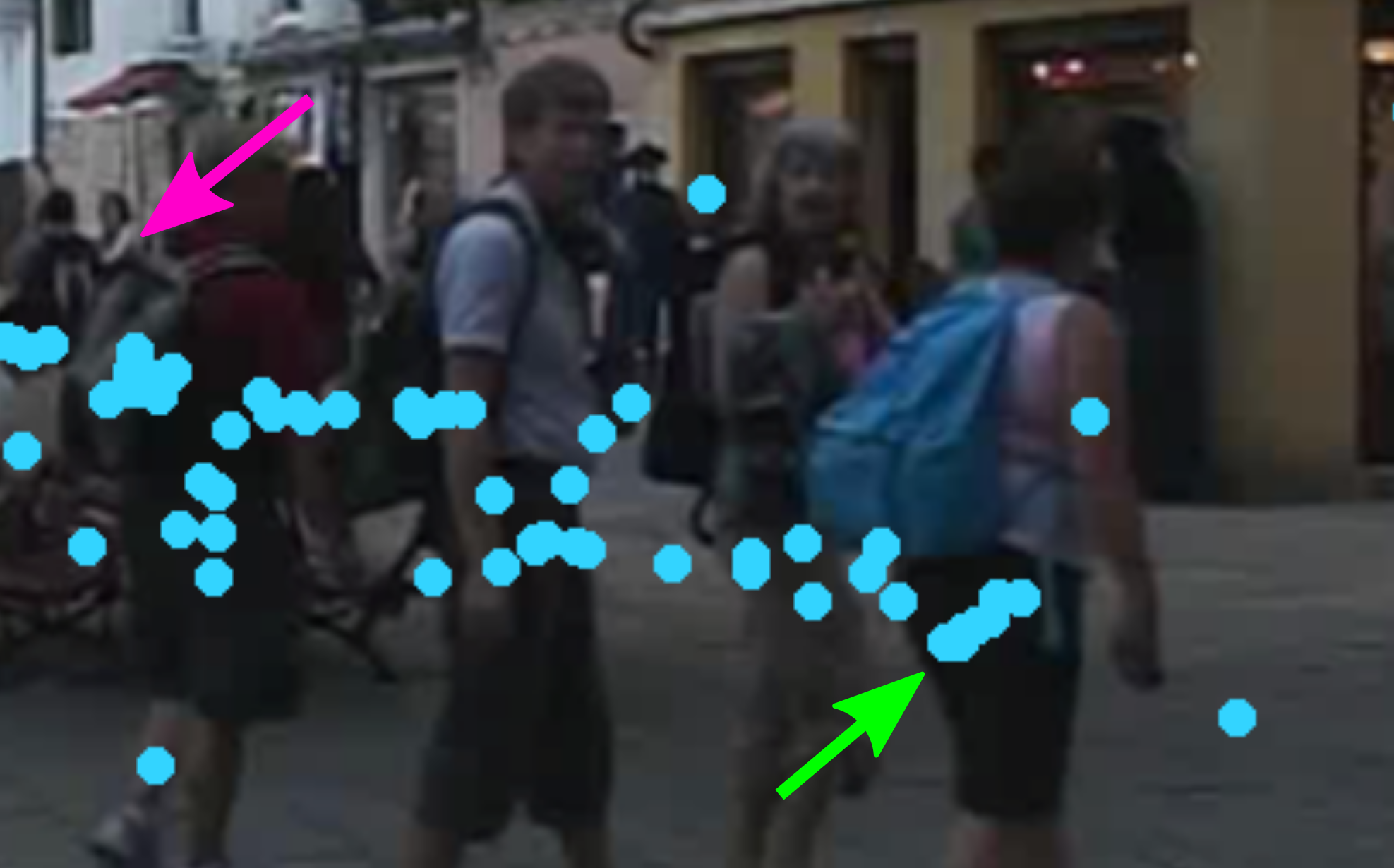}\vspace{-1.5mm}
  \caption{TransTrack~\cite{sun2020transtrack}}
  \label{fig:teaser_transtrack}
\end{subfigure}
\begin{subfigure}{0.24\textwidth}
  \includegraphics[width=\textwidth,height=2.2cm]{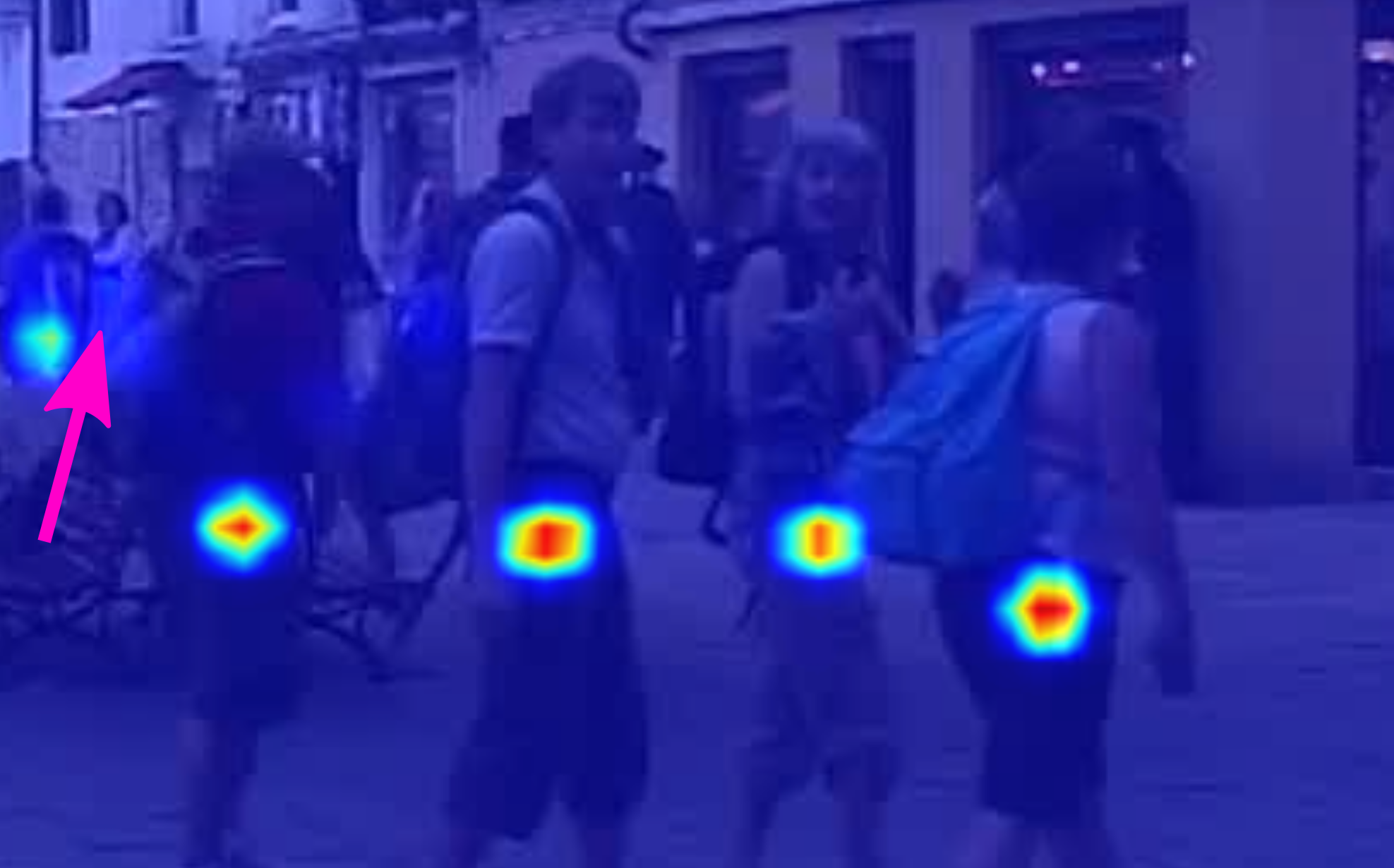}\vspace{-1.5mm}
  \caption{CenterTrack~\cite{zhou2020tracking}}
  \label{fig:teaser_centertrack}
\end{subfigure}\hspace{1mm}%
\begin{subfigure}{0.24\textwidth}
  \includegraphics[width=\textwidth,height=2.2cm]{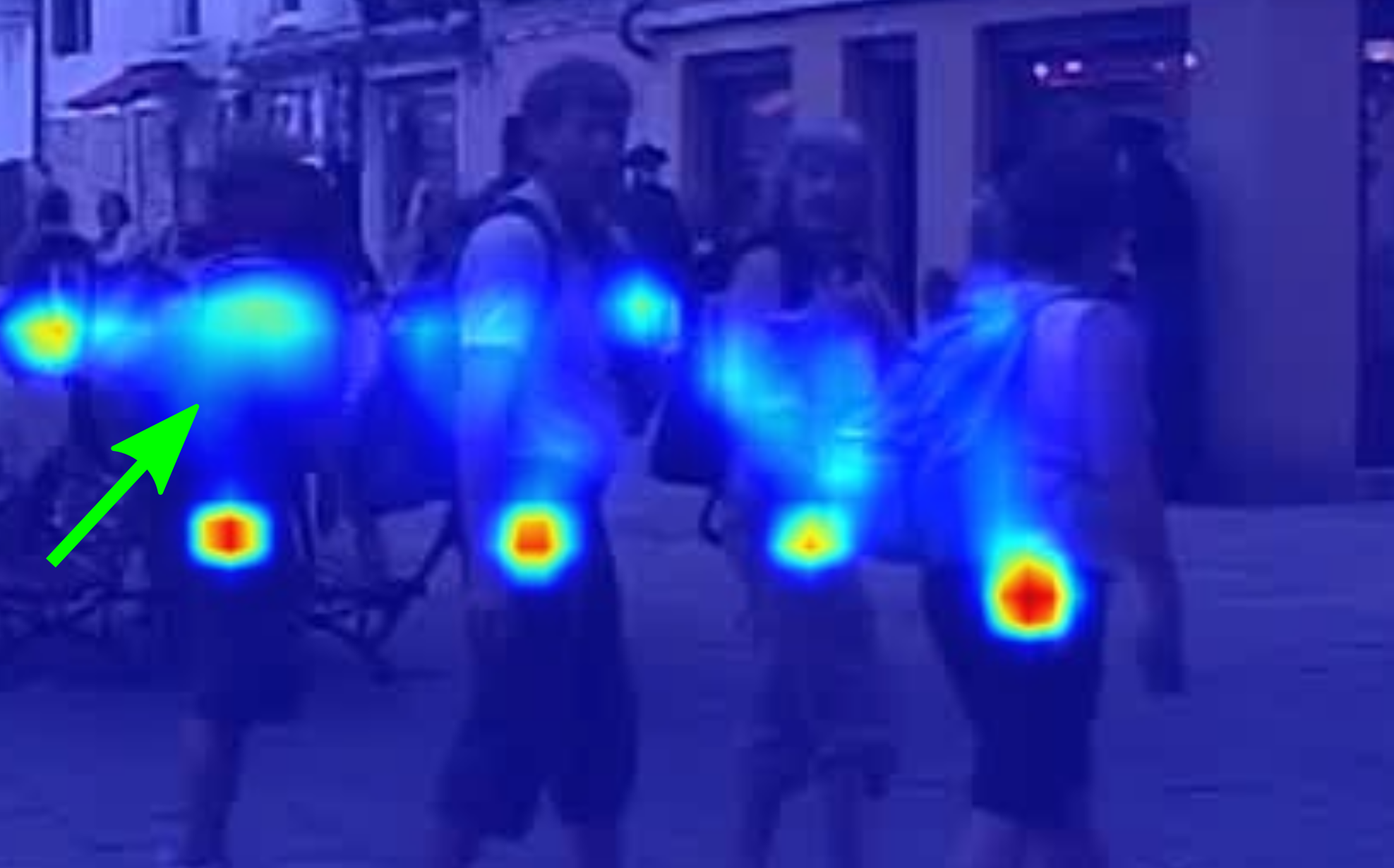}\vspace{-1.5mm}
  \caption{FairMOT~\cite{zhang2020fairmot}}
  \label{fig:teaser_fairmot}
\end{subfigure}
\begin{subfigure}{0.24\textwidth}
  \includegraphics[width=\textwidth,height=2.2cm]{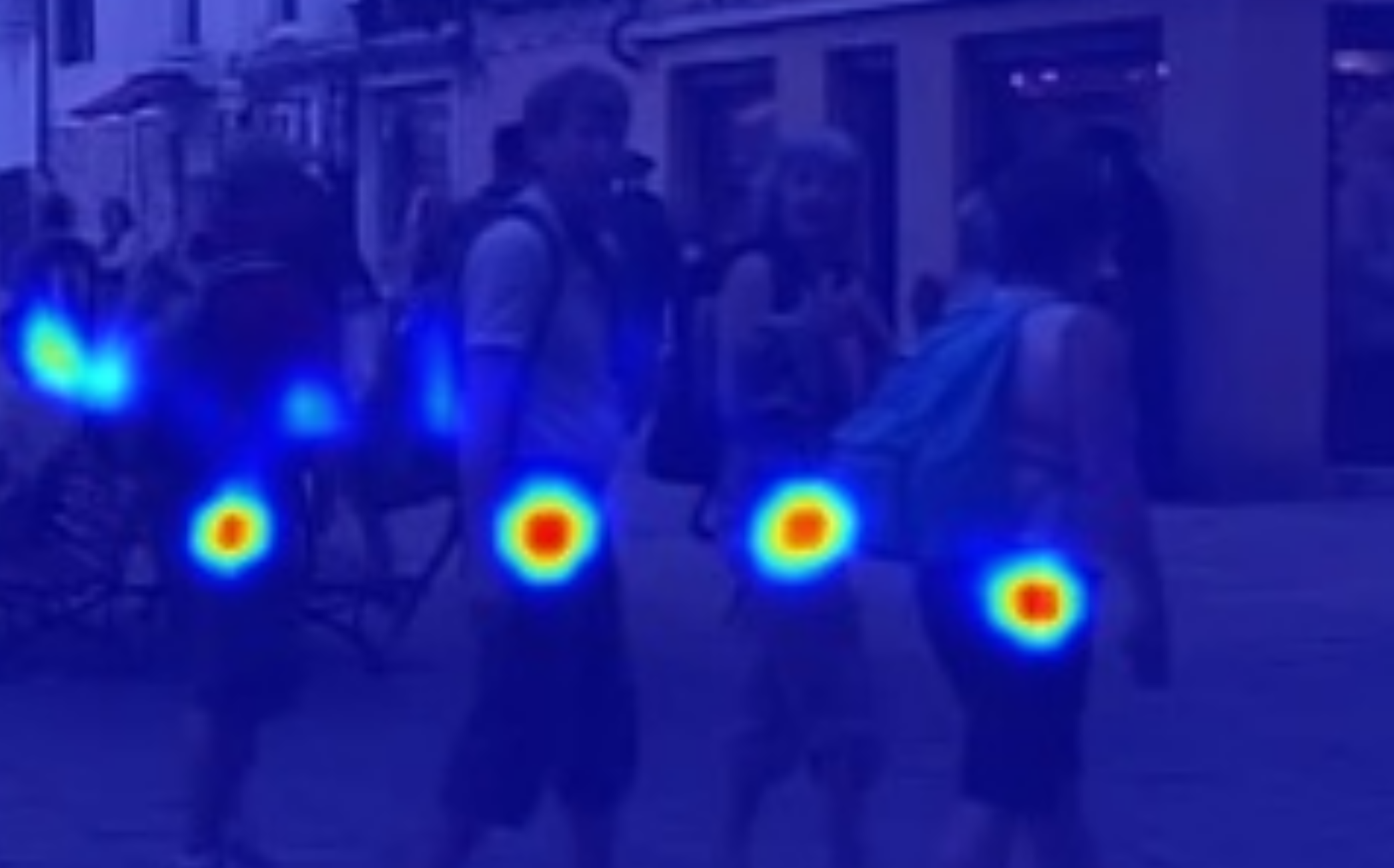}\vspace{-1.5mm}
  \caption{TransCenter}
  \label{fig:teaser_transcenter_b2}
\end{subfigure}
\caption{Detection outputs of state-of-the-art MOT methods: (a) shows the bounding-box centers from the queries in TransTrack~\cite{sun2020transtrack};  (b), (c) are center heatmaps of CenterTrack~\cite{zhou2020tracking}, FairMOT~\cite{zhang2020fairmot}, and (d) is from \method.}\label{fig:teaser}
\end{figure}
\subsection{Center Heatmap Response}\label{subsec:chr}
We qualitatively visualize and compare our center heatmap response in Fig.~\ref{fig:teaser_transcenter_b2} with other two center heatmap-based MOT methods~\cite{zhou2020tracking, zhang2020fairmot} in Fig.~\ref{fig:teaser_fairmot} and Fig.~\ref{fig:teaser_centertrack} as well as the box centers predicted from sparse queries from one concurrent transformer-based MOT method~\cite{sun2020transtrack} in Fig.~\ref{fig:teaser_transtrack}.

Two \emph{concurrent} transformer-based MOT methods~\cite{sun2020transtrack, meinhardt2021trackformer} both use sparse queries, leading to miss detections (pink arrow), that are heavily overlapped, possibly leading to false detections (green arrow). Previous center-based MOT methods~\cite{zhou2020tracking, zhang2020fairmot} suffer from the same problems because the centers are estimated locally.~\method\ is designed to mitigate these two adverse effects by using dense (pixel-level) image-related detection queries producing a dense representation to enable heatmap-based inference and exploiting the attention mechanisms to introduce co-dependency among center predictions.
\begin{figure}[t]
\centering
\captionsetup[subfigure]{labelformat=empty}
\begin{subfigure}{0.23\textwidth}
  \includegraphics[width=\textwidth,height=2.2cm]{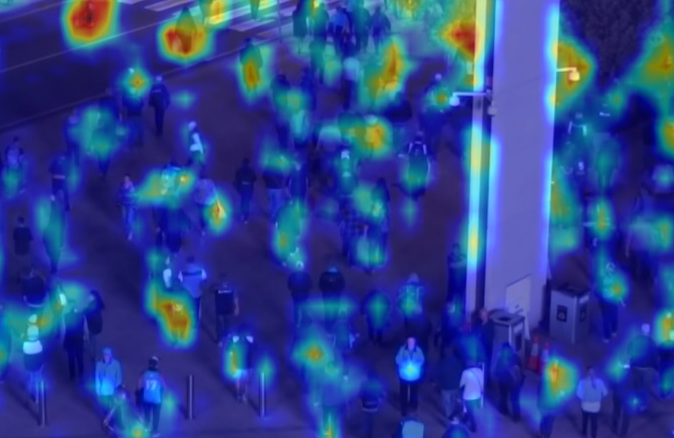}\vspace{-1.5mm}
  \caption{(a)}
  \label{fig:dq_etranscenter-a}
\end{subfigure}\hspace{3mm}%
\captionsetup[subfigure]{labelformat=empty}
\begin{subfigure}{0.23\textwidth}
  \includegraphics[width=\textwidth,height=2.2cm]{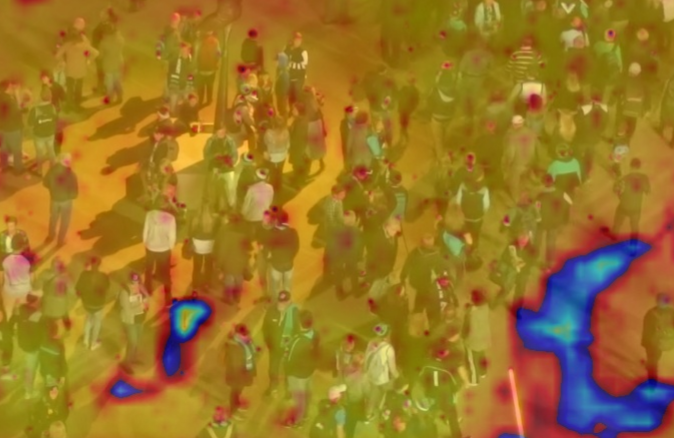}\vspace{-1.5mm}
  \caption{(b)}
  \label{fig:dq_etranscenter-b}
\end{subfigure}
\captionsetup[subfigure]{labelformat=empty}
\begin{subfigure}{0.23\textwidth}
  \includegraphics[width=\textwidth,height=2.2cm]{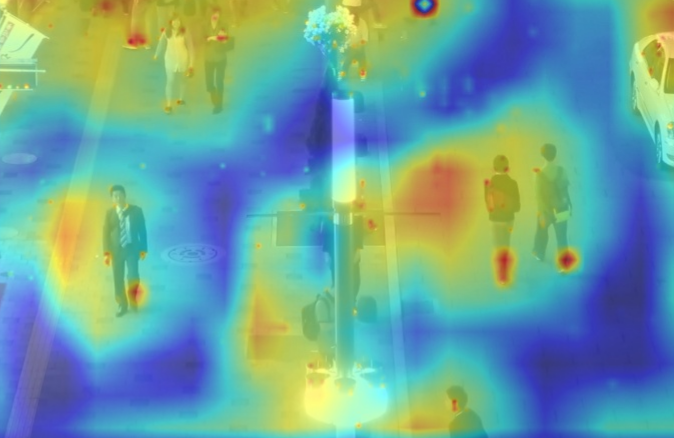}\vspace{-1.5mm}
  \caption{(c)}
  \label{fig:dq_etranscenter-c}
\end{subfigure}\hspace{3mm}%
\captionsetup[subfigure]{labelformat=empty}
\begin{subfigure}{0.23\textwidth}
  \includegraphics[width=\textwidth,height=2.2cm]{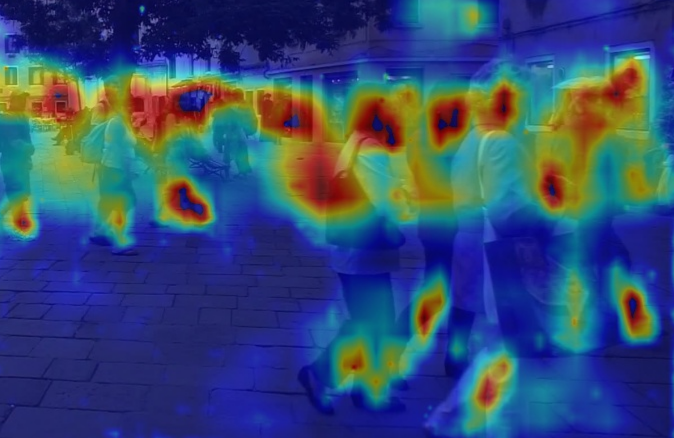}\vspace{-1.5mm}
  \caption{(d)}
  \label{fig:dq_etranscenter-d}
\end{subfigure}
\centering
\caption{Visualization of the detection queries $\mathbf{TQ}$ for \method. The visualization is obtained using the gradient-weighted class activation mapping~\cite{Selvaraju2019}, as detailed in Sec.~\ref{subsec:dqv}. (a) and (b) are from MOT20 while (c) and (d) are from MOT17. Zones with red/orange color represent higher response values and with blue color for lower values.}\label{fig:det_queries_visu}
\end{figure}
\subsection{Detection-Query Visualization}\label{subsec:dqv}
We visualize in Fig.~\ref{fig:det_queries_visu} the detection queries from~\method\ using the gradient-weighted class activation mapping, as described in~\cite{Selvaraju2019}. The intuition is that we can consider the center heatmap prediction as a 2D classification task, similar to image segmentation. The center heatmap response is class-specific where the class is binary with "person" or "background". Precisely, we calculate the gradient from the center heatmap w.r.t.~the detection queries. The gradient is averaged over the image height and width. It is served as weights for the dense detection queries ($\mathbf{DQ}$), producing weighted activations after multiplication. Since our dense detection queries are multi-scale, we re-scale the activations to the input image size and take the average of pixel values from all scales. The activation is plotted above the input image to show the link between their activated values and the object positions. For the visualization purpose, negative values are removed (by a ReLU activation function) and a normalization process by the maximum activation value is performed. From the visualizations, we can see that the detection queries tend to focus on areas where pedestrians are gathered. This is intuitive since the dense detection queries are used to find objects in a given scene.
\begin{figure}[t]
\centering
\captionsetup[subfigure]{labelformat=empty}
\begin{subfigure}{0.23\textwidth}
  \includegraphics[width=\textwidth,height=2.2cm]{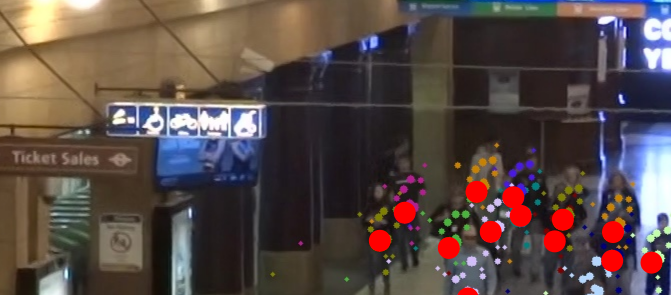}\vspace{-1.5mm}
  \caption{(a)}
  \label{fig:tq_etranscenter-a}
\end{subfigure}\hspace{3mm}%
\captionsetup[subfigure]{labelformat=empty}
\begin{subfigure}{0.23\textwidth}
  \includegraphics[width=\textwidth,height=2.2cm]{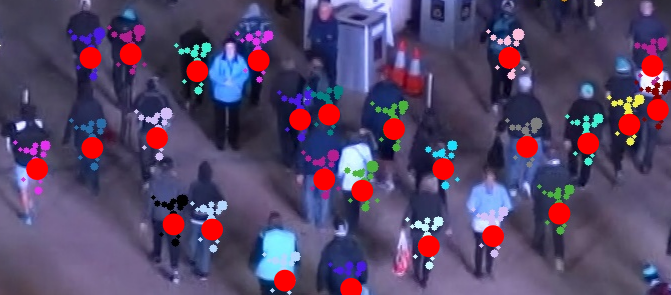}\vspace{-1.5mm}
  \caption{(b)}
  \label{fig:tq_etranscenter-b}
\end{subfigure}
\captionsetup[subfigure]{labelformat=empty}
\begin{subfigure}{0.23\textwidth}
  \includegraphics[width=\textwidth,height=2.2cm]{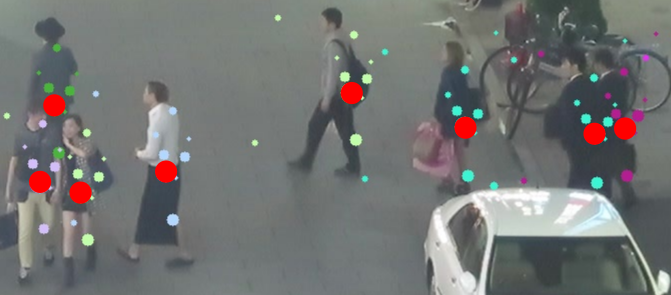}\vspace{-1.5mm}
  \caption{(c)}
  \label{fig:tq_etranscenter-c}
\end{subfigure}\hspace{3mm}%
\captionsetup[subfigure]{labelformat=empty}
\begin{subfigure}{0.23\textwidth}
  \includegraphics[width=\textwidth,height=2.2cm]{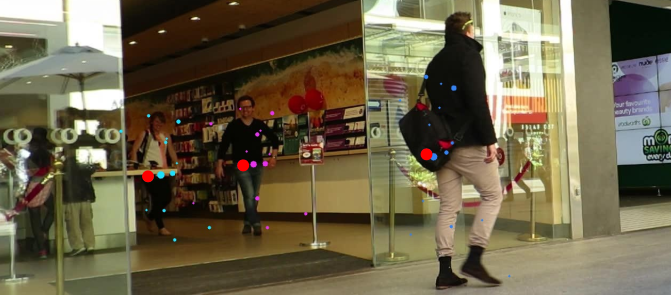}\vspace{-1.5mm}
  \caption{(d)}
  \label{fig:tq_etranscenter-d}
\end{subfigure}
\centering
\caption{Visualization of tracking queries for \method. We visualize the reference points in red dots, and their displacements (circles) and importance (circle radius) with offsets and weights calculated from the tracking queries $\mathbf{TQ}$, as detailed in~ Sec.~\ref{subsec:tqv}. (a) and (b) are from MOT20 while (c) and (d) are from MOT17.}\label{fig:tracking_queries_visu}
\end{figure}\vspace{-5mm}
\subsection{Tracking-Query Visualization}\label{subsec:tqv}
Differently, the tracking queries $\mathbf{TQ}$ in~\method\ are sparse and thus cannot be plotted directly to an image like the detection queries. Therefore, we visualize them differently: Recall that $\mathbf{TQ}$ and $\mathbf{TM}$ interact in TDCA (see Sec.~3.2 in the main paper) for the tracking inside~\method\ Decoder, where we sample object features from the input $\mathbf{TM}$. The sampling locations are obtained by displacing the input reference points (normalized object center positions at the previous frame) with sampling offsets (several offsets per reference point) learned from $\mathbf{TQ}$. Moreover, the sampled features are weighted by the attention weights learned from $\mathbf{TQ}$. Based on this mechanism, we visualize the tracking queries by looking at the sampling locations (displaced reference points) in the image $t$. Concretely, we show in Fig.~\ref{fig:tracking_queries_visu} the reference points (track positions at $t-K$, $K$ is set to 5 for better visualization) as red dots. The sampling locations are shown in circles with different colors referring to different object identities. Their radius is proportional to their attention weight (i.e.~a bigger circle means a higher attention weight) and we filter out sampling locations with weights lower than a threshold (e.g.~0.2) for better visualizations. From Fig.~\ref{fig:tracking_queries_visu}, interestingly, we can see that the sampling locations are the surrounding of the corresponding objects and the weights and the attention weights get smaller when the locations get further from the object. This indicates qualitatively that the tracking queries from the previous time step are searching objects at $t$ in their neighborhood.